\documentclass[10pt]{article} 
\usepackage[accepted]{tmlr}

\usepackage{amsmath,amsfonts,bm}

\def\eqref#1{equation~\ref{#1}}

\def\1{\bm{1}}

\DeclareMathAlphabet{\mathsfit}{\encodingdefault}{\sfdefault}{m}{sl}
\SetMathAlphabet{\mathsfit}{bold}{\encodingdefault}{\sfdefault}{bx}{n}

\DeclareMathOperator*{\argmax}{arg\,max}

\usepackage{hyperref}
\usepackage{url}
\usepackage{booktabs}       
\usepackage{amsfonts}       
\usepackage{nicefrac}       
\usepackage{xspace}
\usepackage{microtype}      
\usepackage{xcolor}         

\usepackage{caption}
\usepackage{subfigure}
\usepackage{hyperref}
\usepackage{wrapfig}
\usepackage{multirow}
\usepackage{amsmath}
\usepackage{amssymb}
\usepackage{mathtools}
\usepackage{amsthm}
\usepackage{enumitem}
\usepackage[flushleft]{threeparttable}
\usepackage{algorithm}
\usepackage{algorithmicx}
\usepackage[noend]{algpseudocode}

\newcommand{\x}{{\boldsymbol{x} }}

\newcommand{\y}{{\bf y}}

\newcommand{\Y}{\mathcal{Y}}

\def\ie{\emph{i.e}\xspace}

\def\1{{\boldmath 1}}
\usepackage[capitalize,noabbrev]{cleveref}

\title{On Pseudo-Labeling for Class-Mismatch Semi-Supervised Learning}

\author{\name Lu Han \email hanlu@lamda.nju.edu.cn \\
      \addr  State Key Laboratory for Novel Software Technology, Nanjing University
      \AND
      \name Han-Jia Ye \email yehj@lamda.nju.edu.cn \\
      \addr State Key Laboratory for Novel Software Technology, Nanjing University
      \AND
      \name De-Chuan Zhan \email zhandc@nju.edu.cn \\
      \addr State Key Laboratory for Novel Software Technology, Nanjing University
      }

\def\month{11}  
\def\year{2022} 
\def\openreview{\url{https://openreview.net/forum?id=tLG26QxoD8}} 

\begin{document}

\maketitle

\begin{abstract}
When there are unlabeled Out-Of-Distribution (OOD) data from other classes, Semi-Supervised Learning (SSL) methods suffer from severe performance degradation and even get worse than merely training on labeled data. 
In this paper, we empirically analyze Pseudo-Labeling (PL) in class-mismatched SSL.
PL is a simple and representative SSL method that transforms SSL problems into supervised learning by creating pseudo-labels for unlabeled data according to the model's prediction. 
We aim to answer two main questions: (1) How do OOD data influence PL? (2) What is the proper usage of OOD data with PL? 
First, we show that the major problem of PL is imbalanced pseudo-labels on OOD data. Second, we find that OOD data can help classify In-Distribution (ID) data given their OOD ground truth labels. 
Based on the findings, we propose to improve PL in class-mismatched SSL with two components -- Re-balanced Pseudo-Labeling (RPL) and Semantic Exploration Clustering (SEC). 
RPL re-balances pseudo-labels of high-confidence data, which simultaneously filters out OOD data and addresses the imbalance problem. SEC uses balanced clustering on low-confidence data to create pseudo-labels on extra classes, simulating the process of training with ground truth. 
Experiments show that our method achieves steady improvement over supervised baseline and state-of-the-art performance under all class mismatch ratios on different benchmarks.
\end{abstract}

\section{Introduction}
\label{sec:intro}

Deep Semi-Supervised Learning (SSL) methods are proposed to reduce dependency on massive labeled data by utilizing a number of cheap, accessible unlabeled data. Pseudo-Labeling~(PL)~\citep{Lee2013Pseudo} is a widely used SSL method. PL is simple yet effective, which creates pseudo-labels according to predictions of the training model itself. Then SSL can be transformed into standard supervised learning. Other representative SSL methods include consistency regularization~\citep{Laine17Temporal,Tarvainen17Mean,Miyato19Virtual}, holistic methods~\citep{Berthelot19Mix,Sohn20Fix}, and generative methods~\citep{Kingma14Semi}. The recent development of SSL shows that these methods have achieved competitive performance to supervised learning methods.

\begin{wrapfigure}{r}[0mm]{0pt}
    \centering
    \includegraphics[width=0.38\linewidth]{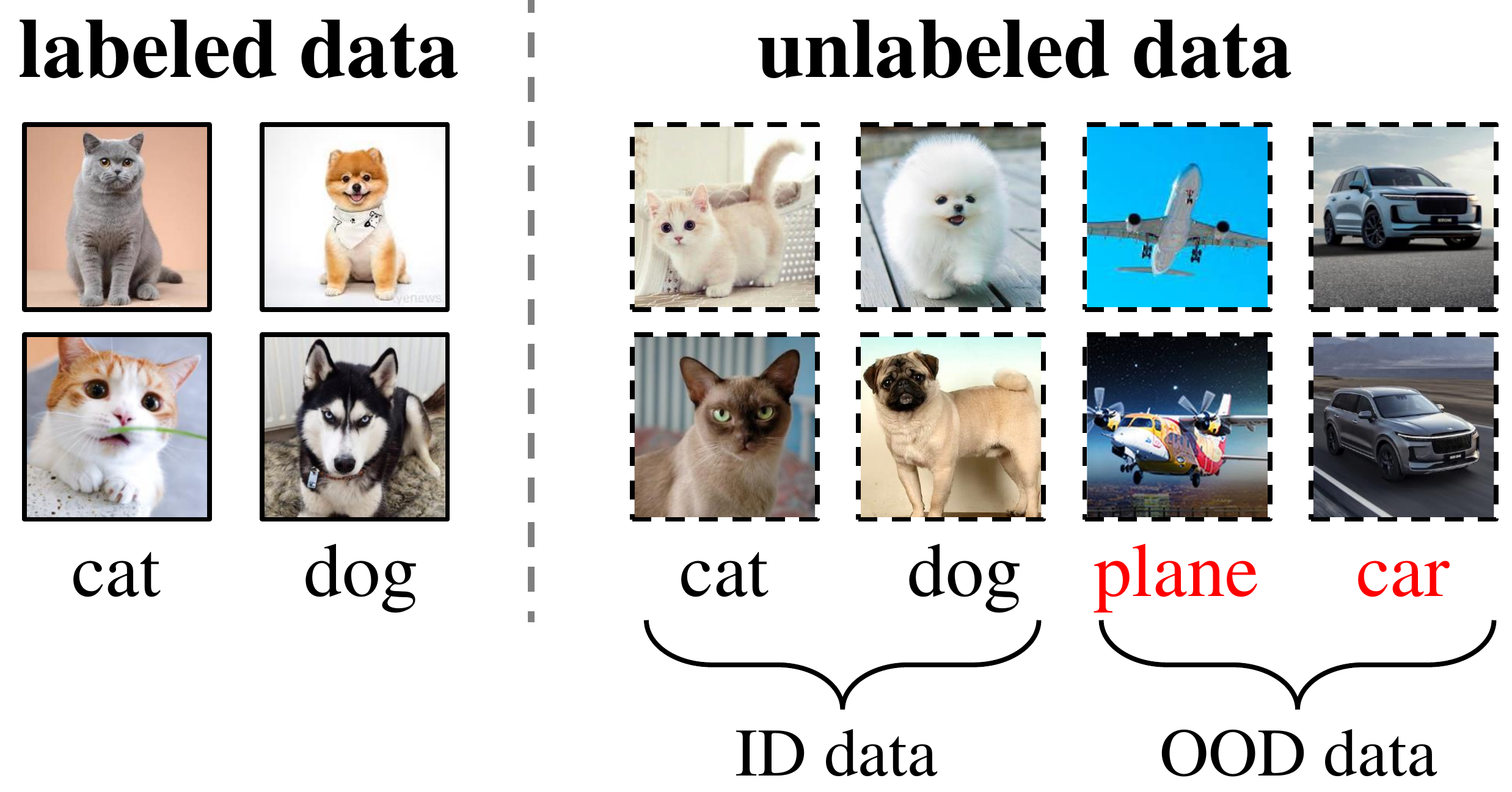}
    \caption{Realistic Semi-Supervised Learning may simultaneously contain unlabeled ID and OOD data. ID data come from the same classes as labeled data while OOD data come from classes that are not seen in labeled data. }
    \vspace{-4mm}
    \label{fig:cm_ssl}
\end{wrapfigure}

However, these SSL methods achieve their good results based on an assumption that unlabeled data are drawn from the same distribution as the labeled data. This assumption can be easily violated in real-world applications. One of the common cases is that some unlabeled data come from {\em unseen classes}. As is illustrated in Figure~\ref{fig:cm_ssl}, in image classification, we can collect a lot of unlabeled images from the internet but usually, they cover broader category concepts than labeled data. \citet{Oliver18Realistic} have shown that under such class-mismatched conditions, the performance of traditional SSL methods is damaged. 
Several methods are proposed for class-mismatched SSL, including filtering out OOD data~\citep{Yu20Multi,Chen20Semi}, down weighting OOD data~\citep{Chen20Semi}, and re-using OOD data by neural style transfer~\citep{Luo21On}/self-supervised learning~\citep{Huang21Trash}. Although these methods achieve good results, why OOD data damage performance and how will OOD data help remain unclear.

In this paper, we focus on empirically analyzing one representative family of the SSL method --- PL in class-mismatched SSL and give some answers to these two questions.
(1) How do OOD data influence PL? (2) What are suitable pseudo-labels for OOD data? For question (1), we investigate pseudo-labels created by PL. The main finding is that pseudo-labels on OOD data tend to be imbalanced while on ID data, they remain balanced. We further show that PL's performance is damaged due to such an imbalance in OOD data. For question (2), several strategies for labeling OOD data are investigated. We conclude that it is beneficial when labeling OOD data as a class different from ID data, and the performance can be further improved when pseudo-labels partition unlabeled OOD data into their semantic clusters.

Based on the experimental analyses, we propose a two-branched model called $\Upsilon$-Model, which processes unlabeled data according to their confidence score on ID classes.
The first branch performs Re-balanced Pseudo-Labeling~(RPL) on high-confidence data. 
It utilizes the property of imbalanced pseudo-labels on OOD data, truncating the number of pseudo-labeled data for each class to their minimum. This procedure filters out many OOD data and also prevents the negative effect of imbalanced pseudo-labels. For the other branch, Semantic Exploration Clustering~(SEC) is performed on low-confidence data. They are considered OOD data and their semantics will be mined by clustering into different partitions on extra classes. The clustering result provides better pseudo-labels for these OOD data than vanilla PL.
Experiments on different SSL benchmarks show that our model can achieve steady improvement in comparison to the supervised baseline. Our contributions are:


\begin{itemize}[leftmargin=4mm,itemsep=0mm]
    \vspace{-1mm}
    \item We analyze PL for ID and OOD data. The findings lead to two primary conclusions: (1) Imbalance of pseudo-labels on OOD data damages PL's performance. (2) Best pseudo-labels for unlabeled OOD data are those different from ID classes and partitioning them into their semantic clusters. 
    \item We propose our two-branched $\Upsilon$-Model. One branch re-balances pseudo-labels on ID classes and filters out OOD data. The other branch explores the semantics of OOD data by clustering on extra classes.
    \item Experiments on different SSL benchmarks empirically validate the effectiveness of our model.
    \vspace{-1mm}
\end{itemize}

\section{Preliminary}
\subsection{Class-Mismatched SSL}
Similar to the SSL problem, the training dataset of the class-mismatched SSL problem contains $n$ ID labeled samples $\mathcal{D}_l = \{(\x_{li},y_{li})\}^n_{i=1}$ and $m$ unlabeled samples $\mathcal{D}_u =\{\x_{ui}\}^m_{i=1}$, (usually, $m\gg n$,) $y_{li} \in \mathcal{Y}_{ID} = \{1,\ldots,K_{ID}\}$, while different from SSL, the underlying ground truth $\y_u$ of unlabeled data may be different from labeled data. \ie,  $y_{uj} \in \mathcal{Y}_{ID} \cup \mathcal{Y}_{OOD}, \mathcal{Y}_{OOD} = \{K_{ID}+1,\ldots, K_{ID}+K_{OOD}\}$. The goal of class-mismatched SSL is to \textbf{correctly classify ID samples into} $
\Y_{ID}$ using labeled set with ID samples and unlabeled set possibly with OOD samples.

\subsection{Pseudo-Labeling}
{\it Pseudo-Labeling} (PL) leverages the idea that we can use the model itself to obtain artificial labels for unlabeled data~\citep{Lee2013Pseudo}. PL first performs supervised learning on labeled data to get a pre-trained model $f$, which outputs the probability of belonging to each ID class. 
Given $c(\x)$ is the confidence score for $
\x$
\begin{equation}
    c(\x) = \max_{y \in \mathcal{Y}_{ID}} f(y|\x) ,
    \label{eq:confidence}
\end{equation}
PL creates the pseudo-labels for each unlabeled sample:
\begin{equation}
    y' = \begin{cases}
        \argmax_{y \in \mathcal{Y}_{ID}}f(y|\x)&,\quad c(\x) > \tau \\
        \text{reject} &,\quad  \text{otherwise}
    \end{cases} ,
    \label{eq:pl}
\end{equation}
All pseudo-labeled unlabeled data will be treated as labeled data for the next supervised learning generation. PL iteratively performs supervised learning and pseudo-label creation until stops. 
\section{Analysis of Pseudo-Labeling in Class-Mismatched SSL}
\label{sec:analysis}

\begin{figure}[htp]
	\centering
	\setcounter{subfigure}{0}
	\subfigure[Pseudo-labels on ID]{
		\label{fig:subfig:id_distr} 
		\includegraphics[width=0.30\linewidth]{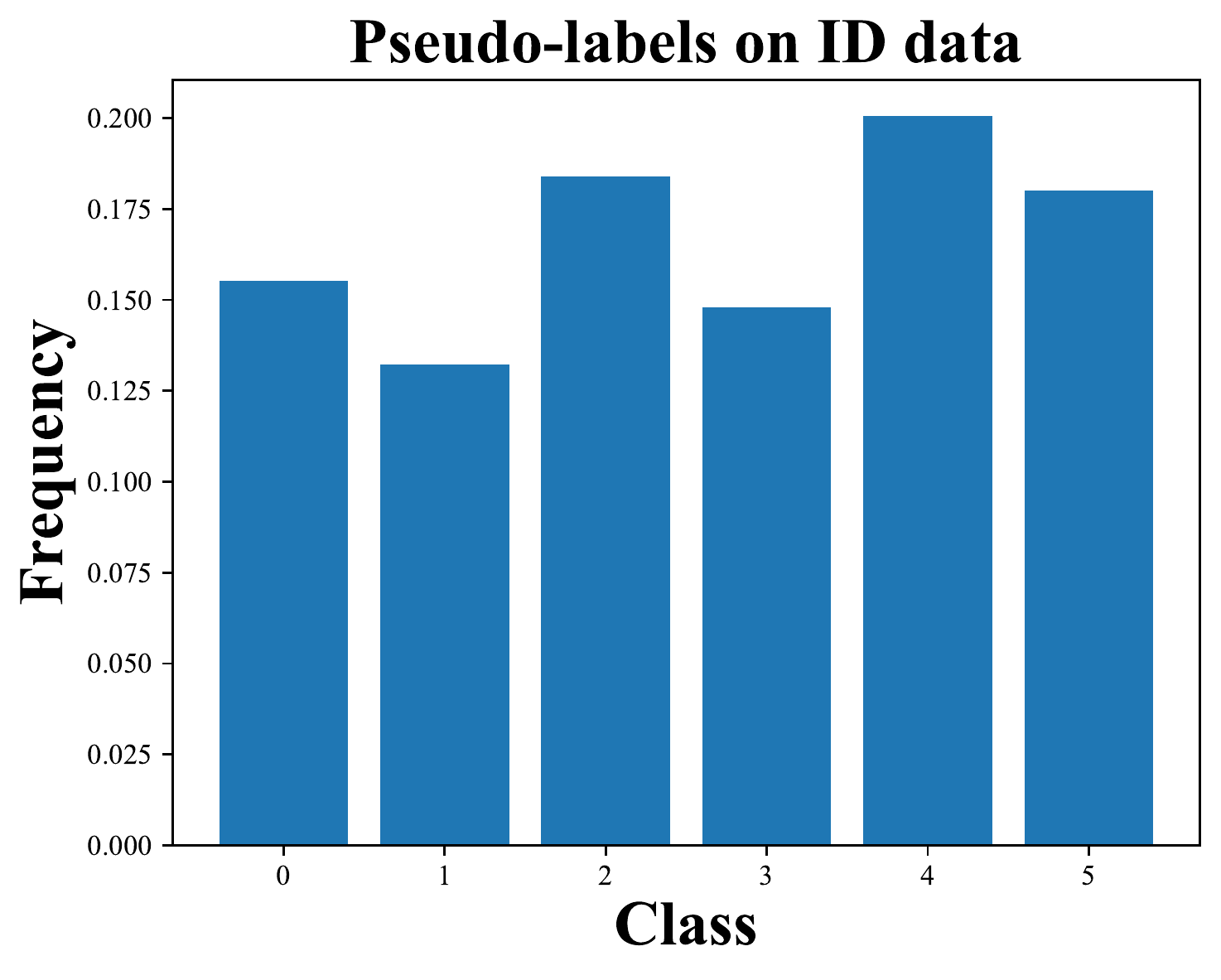}
	}\hspace{-1mm}
	\subfigure[Pseudo-labels on OOD]{
		\label{fig:subfig:ood_distr} 
		\includegraphics[width=0.30\linewidth]{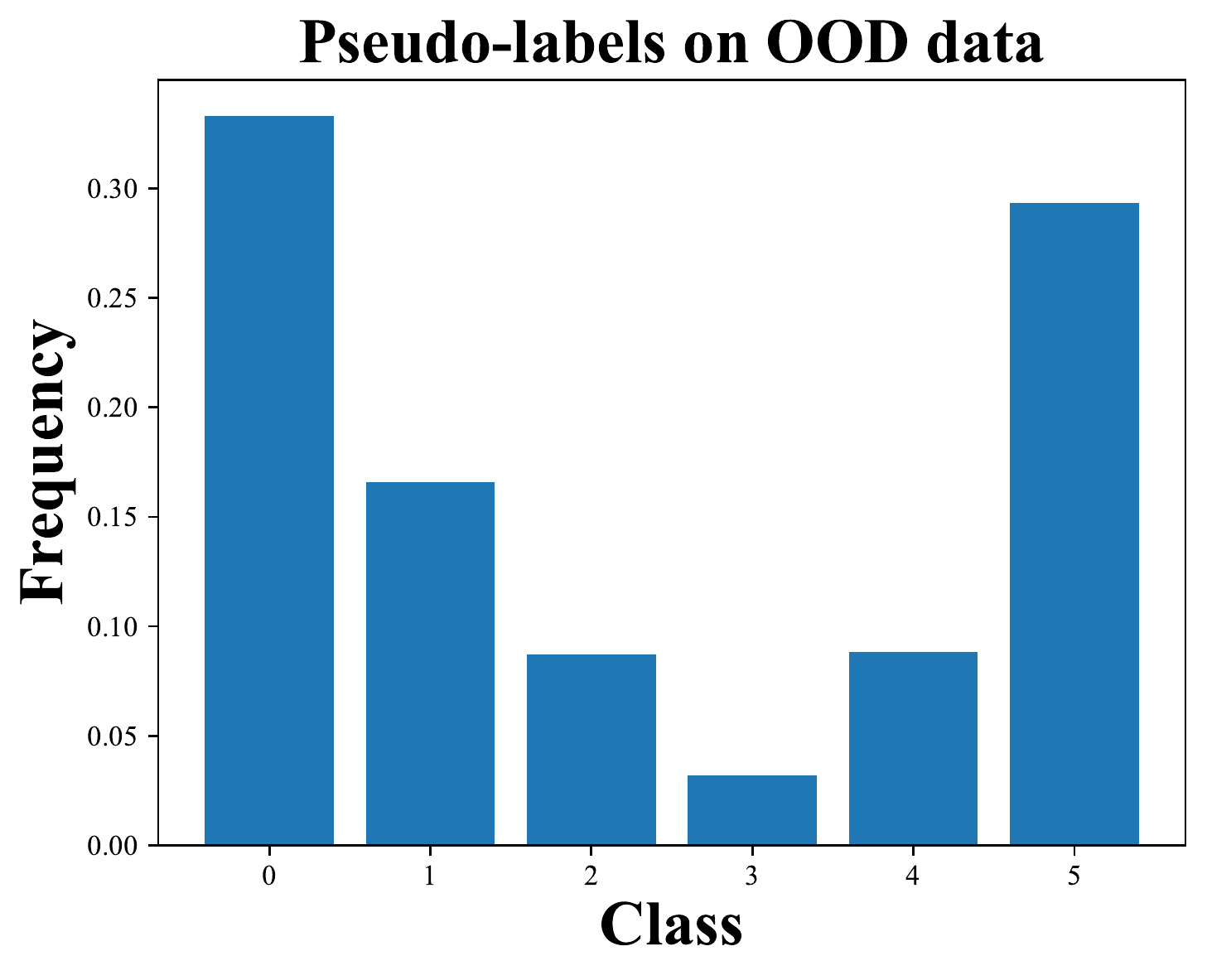}
	}
	\hspace{-1mm}
	\subfigure[Imbalance ratio]{
		\label{fig:subfig:id_ood} 
		\includegraphics[width=0.30\linewidth]{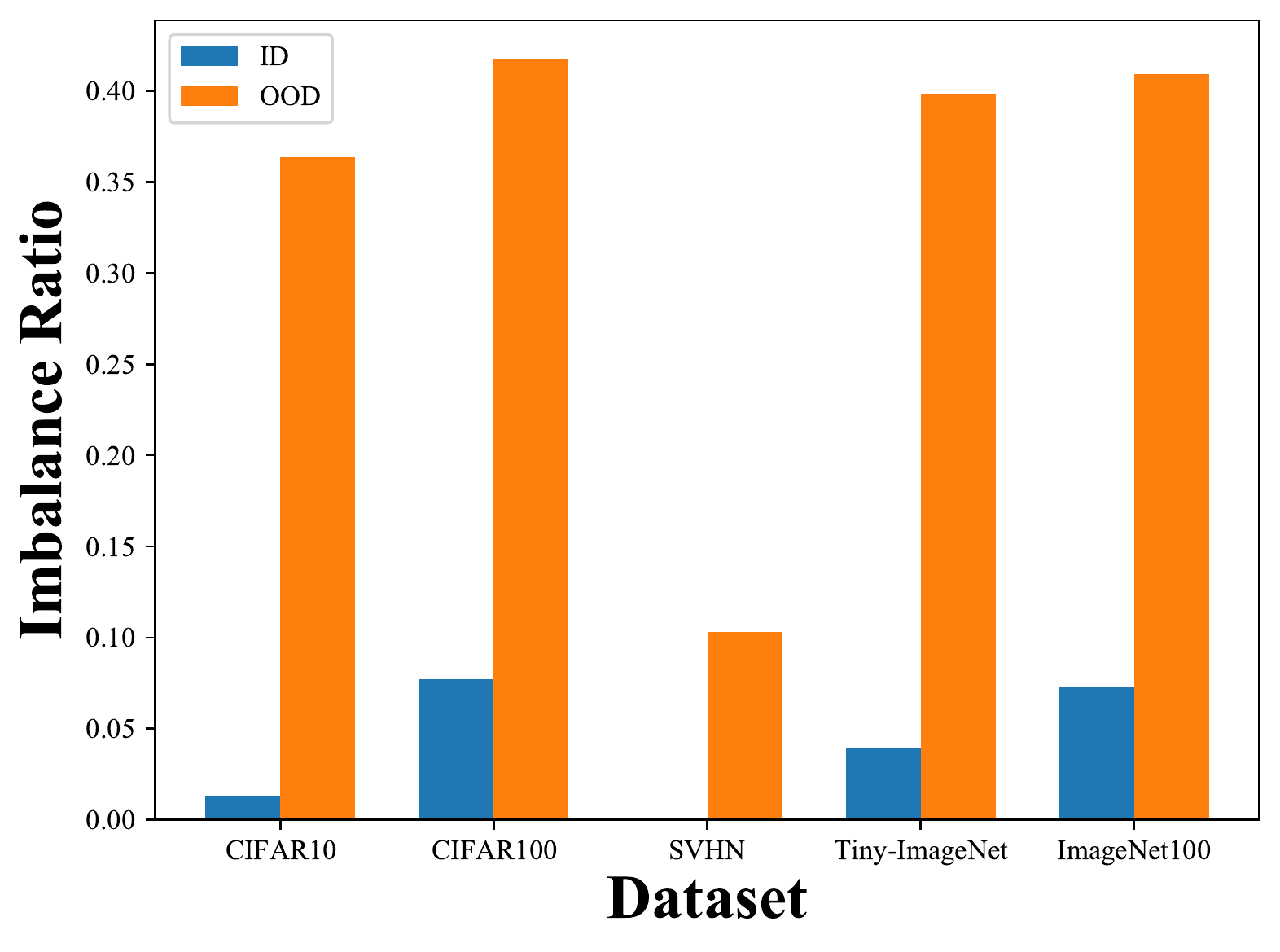}
	}
	\vspace{-3mm}
	\caption{Analysis of the pseudo-label distributions created by the pre-trained model. 
		(a) On ID data, pseudo-label distribution is balanced since they share the same distribution with the labeled data. (b) On OOD data, the pseudo-label distribution is {\em imbalanced}. (c) The imbalance ratio of pseudo-labels on ID and OOD data on other datasets.
		The imbalance ratio is computed by KL-divergence with uniform distribution.}
	\label{fig:pred_distr} 
\end{figure}

In class-mismatched SSL, vanilla PL can only create pseudo-labels on ID classes even for OOD data. We will analyze how these OOD data influence vanilla PL and what are the better pseudo-labels for them in this section.
Experiments are carried out on totally five kinds of datasets. (We use ($n$/$m$) to represent $n$ ID classes and $m$ OOD classes.)
\begin{itemize}[leftmargin=8mm,itemsep=0mm]
	\vspace{-1mm}
	\item \textbf{CIFAR10 (6/4)} : created from \textbf{CIFAR10}~\citep{Krizhevsky2009Learning}.It takes the 6 animal classes as ID classes and 4 vehicle classes as OOD classes. We select 400 labeled samples for each ID class and totally 20,000 unlabeled samples from ID and OOD classes.
	\item \textbf{SVHN (6/4)}: We select the first ``0''-``5'' as ID classes and the rest as OOD. We select 100 labeled samples for each ID class and totally 20,000 unlabeled samples.
	\item \textbf{CIFAR100 (50/50)}: created from \textbf{CIFAR100}~\citep{Krizhevsky2009Learning}. The first 50 classes are taken as ID classes and the rest as OOD classes. We select 100 labeled samples for each ID class and a total of 20,000 unlabeled samples.
	\item \textbf{Tiny ImageNet (100/100)}: created from \textbf{Tiny ImageNet}, which is a subset of \textbf{ImageNet}~\citep{Deng09ImageNet} with images downscaled to 64 $\times$ 64 from 200 classes. The first 100 classes are taken as ID classes and the rest as OOD classes. We select 100 labeled samples for each ID class and 40,000 unlabeled samples.
	\item \textbf{ImageNet100 (50/50)}: created from the 100 class subset of ImageNet~\citep{Deng09ImageNet}. The first 50 classes are taken as ID classes and the rest as OOD classes. We select 100 labeled samples for each ID class and a total of 20,000 unlabeled samples.
\end{itemize} 
\textbf{Here we use C to represent CIFAR, TIN to represent Tiny ImageNet, IN to represent ImageNet for short}. We vary the ratio of unlabeled images to modulate class distribution mismatch. For example, the extent is 50\% means half of the unlabeled data comes from ID classes and the others come from OOD classes.
We use Wide-ResNet-28-2~\citep{Zagoruyko16Wide} as our backbone. We also adopt data augmentation techniques including random resized crop, random color distortion and random horizontal flip. 
For each epoch, we iterate over the unlabeled set and random sample labeled data, each unlabeled and labeled mini-batch contains 128 samples. We adopt Adam as the
optimization algorithm with the initial learning rate $3\times 10^{-3}$ and train for 400 epochs. Averaged accuracies of the last 20 epochs are reported, pretending there is no reliable (too small) validation set to perform early stop~\citep{Oliver18Realistic}.

\subsection{Imbalance of Pseudo-labels on OOD Data}
In this section, we analyze the \textbf{pre-trained model} that creates the first set of pseudo-labels, and the \textbf{final model} trained by Pseudo-Labeling. 

\vspace{-3mm}
\paragraph{Pre-trained model.} Like what is concluded in OOD detection~\citep{Hendrycks17Baseline}, ID data tend to have higher confidence than OOD data, but there are still considerable OOD data with high confidence. In class-mismatched SSL, the unlabeled data are in much larger quantities. When the class mismatch ratio is large, there are quite a few OOD data with high confidence scores. We will show in the final model experiments that these high-confidence OOD data damage performance. Secondly, we study pseudo-labels on both ID data and OOD data. \cref{fig:subfig:id_distr} shows that pseudo-labels ID data is balanced. However, they are rather imbalanced on OOD data ~(\cref{fig:subfig:ood_distr}). 
Such difference in created pseudo-labels is attributed to the different distribution they are drawn from. Samples with certain pattern bias to certain classes. ID data bias to ID classes uniformly because they are sampled by the same distribution. However, with little probability, OOD data will also bias to ID classes uniformly since they have little relevance to ID data.\footnote{ How OOD data are generated affects the imbalance ratio. We experiment on different OOD class settings in~\cref{app_subsec:imbalance_ood} to show that imbalanced pseudo-labels on OOD data is a general phenomenon for non-curated natural datasets.}

\begin{figure}[htp]
	\centering
	\setcounter{subfigure}{0}
	\subfigure[Performance of PL]{
		\label{fig:subfig:pl_ratio}
		\includegraphics[width=0.38\linewidth]{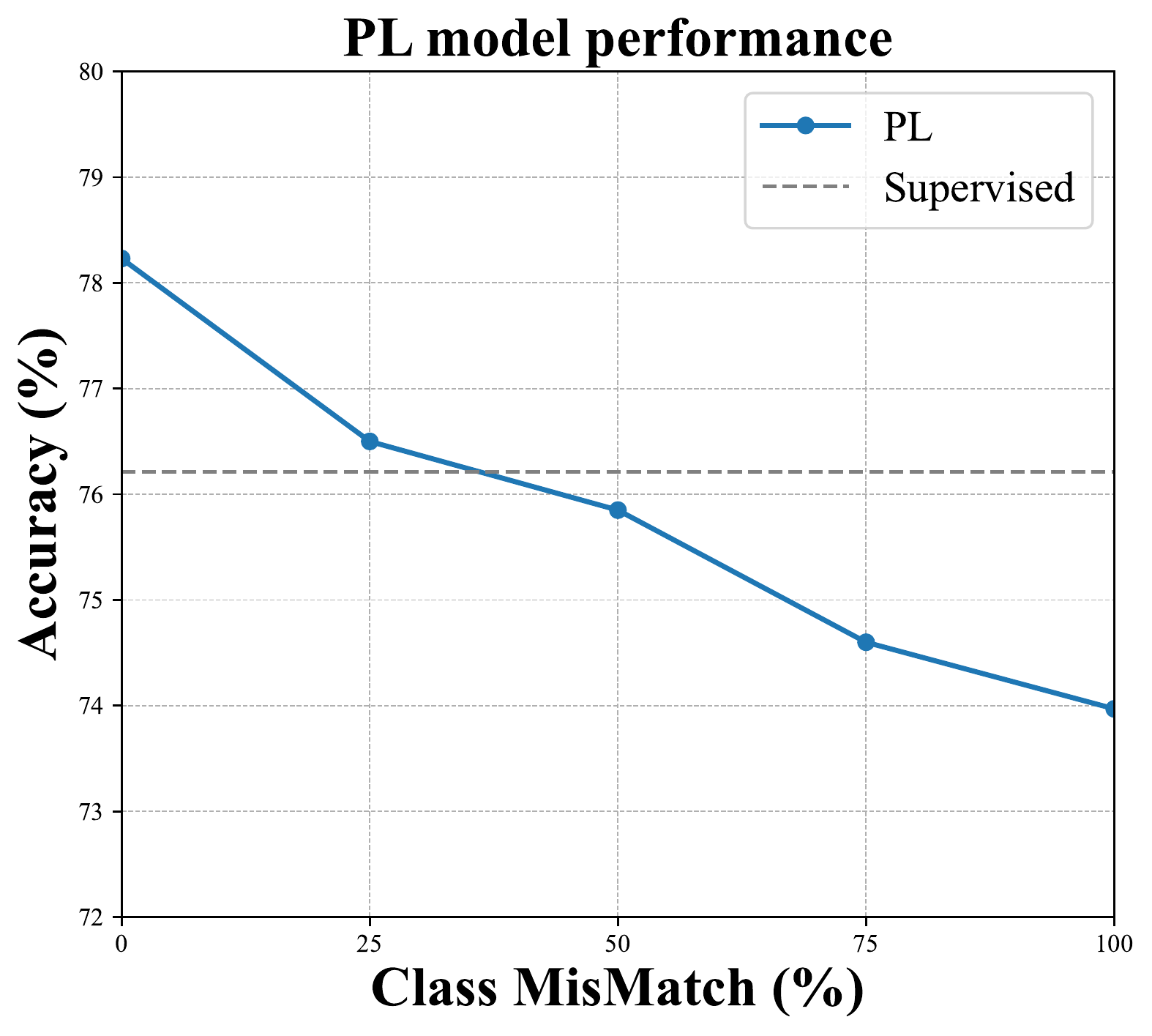}
		
	}
	\hspace{-4mm}
	\subfigure[Confusion Matrix of PL model]{
		\label{fig:subfig:cm_pl} 
		\includegraphics[width=0.56\linewidth]{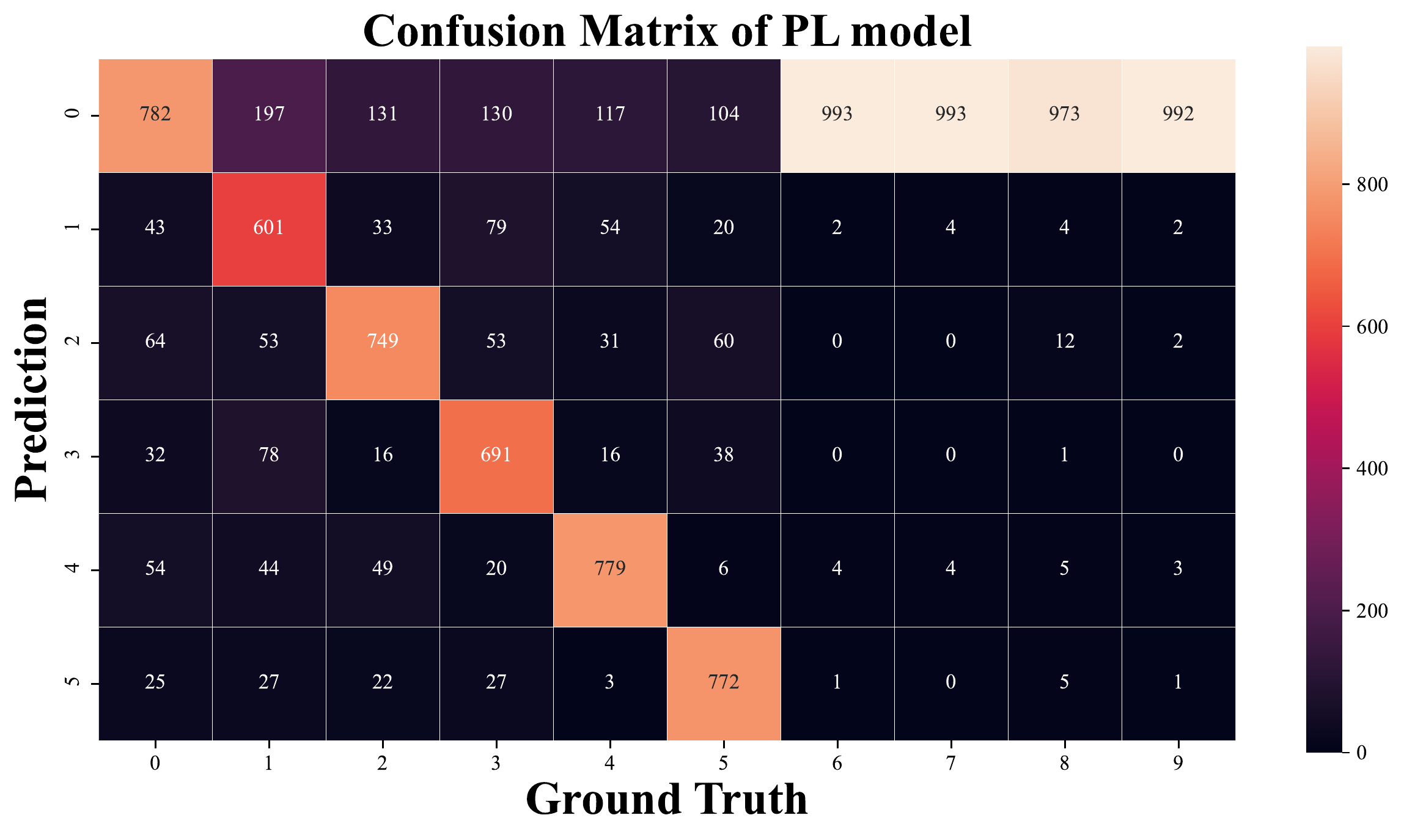}
	}
	\vspace{-3mm}
	\caption{(a) PL model degrades as the mismatch ratio increases. (b) Confusion matrix of PL model when the mismatch ratio = $100\%$. It demonstrates that the imbalance of pseudo labels on OOD data affects PL's performance. A lot of ID samples with class 1-5 are misclassified into class 0. Also, as the PL process continues, the imbalance of pseudo-labels on OOD data get even worse.}
	\vspace{-3mm}
\end{figure}

\vspace{-2mm}
\paragraph{Final Pseudo-Labeling model.} As an old saying goes, a good beginning is half done. However, such imbalance of the first set of pseudo-labels starts PL model badly when there is a large portion of OOD data, putting the model in danger of imbalanced learning. We run vanilla PL and show that the \textit{imbalance of pseudo-labels harms the performance.} Figure~\ref{fig:subfig:pl_ratio} shows the performance of PL model with different OOD ratios. In accord with~\citep{Oliver18Realistic}, PL model degrades as the portion of OOD data gets larger. Figure~\ref{fig:subfig:cm_pl} displays the confusion matrix of the PL model on the whole test set containing both ID and OOD data. Since only 6 classes are known to us, the confusion matrix is a rectangle. 
We can see almost all the OOD samples (class 6-9) are classified as class 0, which means the imbalance effect on OOD data gets even worse as the PL training goes on. The possible reason is that, unlike Pseudo-Labeling on ID data, supervision of labeled data can not help correct pseudo-labels on OOD data. Thus the imbalance continuously deteriorates. The imbalance on OOD data also influences classification performance on ID data. Samples of major classes (class 0) overwhelm the loss and gradient, leading to a degenerate model~\citep{Lin17Focal}. We can see the PL model mistakenly classifies many of data with class 1-5 into class 0. 

\subsection{Pseudo-Labeling Strategy for OOD data}
\label{subsec:pl_ood}

The previous section shows OOD data hurt the performance of vanilla PL. Then here comes the question: Assuming that we already know which data are OOD, \textbf{how do we use these OOD data? Is omitting them the best way? If not, what are the better pseudo-labels for them?} To answer these questions, we investigate four strategies to create pseudo-labels for OOD data:
\begin{itemize}[leftmargin=8mm,itemsep=0mm]
    \vspace{-1mm}
    \item \textbf{Baseline.} This baseline omits all the OOD data and only trains on the labeled ID data.
    \item \textbf{Re-Assigned Labeling\footnote{Note that Re-Assigned Labeling has many possibilities. If there are $n$ ID classes and $m$ OOD classes, $A_n^m$ possible assignments exist. It is impossible to experiment on all of them. To deal with it, we randomly choose $10$ possible assignment and pick the maximum performance among them.}.} This strategy assigns data of each OOD class to an ID class. It ensures that different OOD class is assigned to different ID class, keeping the semantics unchanged between OOD classes. For example, (ship, trunk, airline,  automobile) can be assigned to (bird, cat, deer, dog). This strategy can be seen as training a classifier of ``super-classes''. 
    \item \textbf{Open-Set Labeling.} This strategy is named after the related setting -- Open-Set Recognition~\citep{Scheirer13Toward,Bendale16Towards}. This strategy treats all OOD data as one unified class $K_{ID}+1$. Thus this model outputs probability over $K_{ID}+1$ classes.
    \item \textbf{Oracle Labeling.} This strategy uses the ground truth of OOD data. Thus this model outputs probability over $K_{ID}+K_{OOD}$ classes.
    \vspace{-1mm}
\end{itemize}

\begin{figure}[tp]
    \centering
    \includegraphics[width=\linewidth]{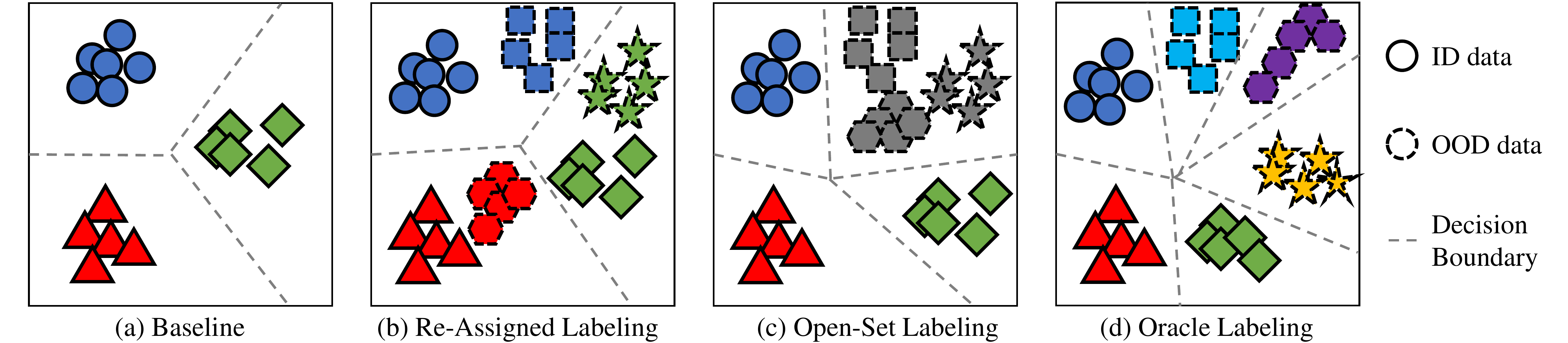}
    \caption{Four strategies of how to label OOD data. Different shapes represent different ground truths. Data with the same color are labeled as the same classes. A shape with a solid outline means it is an ID sample while OOD data are represented with a dashed line. (a) No Labeling acts as a baseline where OOD data are omitted. (b) Re-Assigned Labeling re-labels OOD data to certain ID classes. (c) Open-Set Labeling labels all the OOD data as a unified class. (d) Oracle Labeling uses the ground truths of OOD data.
    }
    \vspace{-4mm}
    \label{fig:ces}
\end{figure}

Note that Open-Set Labeling and Oracle Labeling can classify samples into more than $K_{ID}$ classes. However, during evaluation, we only classify samples into $K_{ID}$ ID classes. For these models, the predicted label $\hat{y}$ of a test sample $\x$ is calculated as:
\begin{equation}
    \hat{y}(x) = \argmax_{y \in \mathcal{Y}_{ID}} f(y|\x)
    \label{eq:pred}
\end{equation}
the overall comparison of the four strategies is illustrated in Figure~\ref{fig:ces}. We also report test accuracy on the five datasets when the class-mismatched ratio is $100\%$ in \cref{tb:appendix_pl_ood}. From the results, we can get several important conclusions. (1) Re-Assigned Labeling underperforms baseline a little. This indicates that assigning samples with OOD classes to ID classes does not help the model distinguish between ID classes even if we somehow know which OOD data are semantically different. It also reveals that performing vanilla PL on OOD data may never help even if we do it perfectly. (2) Open-Set Labeling outperforms baseline, which indicates it improves the performance if we label the OOD data as a class other than ID classes. (3) We can see Oracle Labeling improves performance and achieves the best results among the four strategies. It means that in addition to labeling OOD data as extra classes, if we can further assign OOD data with different semantics to different classes, the model will achieve better results.


\begin{table}[htp]
	\caption{Performance of five different pseudo-labeling strategies on different datasets. It can be concluded: (1) Re-Assigned Labeling underperforms baseline. (2) Open-Set Labeling outperforms baseline a little. (3) Oracle Labeling improves performance and achieves the best results.}
	\centering
	\begin{tabular}{lccccc}
		\toprule
		& \textbf{C10 (6/4)}   & \textbf{SVHN (6/4)}   & \textbf{C100 (50/50)} & \textbf{TIN (100/100)} & \textbf{IN (50/50)}\\
		\midrule
		Baseline             & 76.21 (-)                 & 88.33 (-)        & 58.68 (-)            & 39.08 (-)            & 48.12 (-)     \\
		Re-Assigned & 75.99(-0.22)    & 84.40({\color{red}-3.93}) & 50.52({\color{red}-8.16})     & 34.90({\color{red}-4.76})  &      45.60 ({\color{red}-2.52})     \\
		Open-Set    & 77.95({\color{teal}+1.74})   & 88.31(-0.02) & 58.76({\color{teal}+0.08})     & 40.06({\color{teal}+0.86})   &   49.56({\color{teal}+1.44})       \\
		Oracle      & 79.25({\color{teal} +3.04})    & 92.07({\color{teal}+3.74}) & 63.90({\color{teal}+3.73})     & 45.28({\color{teal}+6.78})   &      55.96({\color{teal}+7.84}) \\
		\bottomrule
	\end{tabular}
	\label{tb:appendix_pl_ood}
	\vspace{-2mm}
\end{table}
\vspace{-2mm}
\paragraph{Discussion.} Why does Oracle Labeling consistently outperform Open-Set Labeling by a large margin? We think the most important reason is that Oracle Labeling utilizes information among OOD classes. For example, in the experiment of CIFAR10(6/4), open-set labeling takes all the vehicle samples as one class, ignoring the fact that they come from four categories -- airplane, automobile, ship and truck. The difference among data of these classes may provide useful information not covered by labeled data of ID classes, especially when the labeled data is not plenty. For example, learning to distinguish between airplane and truck helps distinguish between bird and dog by judging whether having wings or not. But the model trained by open-set labeling loses such benefit. In contrast, Oracle labeling can help the model capture this information by utilizing the ground truth labels. Consequently, oracle labeling performs better than open-set labeling.


\subsection{Summary of Section}
\label{subsec:summary}
In this section, we study the behavior of the Pseudo-Labeling model in class-mismatched SSL. We summarize several important conclusions here:
\begin{enumerate}[itemindent=0pt,itemsep=1pt,leftmargin=28mm,label=Conclusion \arabic*: ,ref=\arabic*]
    \item\label{cc:ce_imbalance}Classification model trained with labeled ID data creates imbalanced pseudo-labels on OOD data while balanced pseudo-labels on ID data. 
    \item\label{cc:pl_imbalance}The vanilla PL  makes the imbalance of pseudo-labels deteriorate, damaging the classification performance on ID data. 
    \item \label{cc:ood_label_id}Labeling OOD data as ID classes does not help and may even perform worse. 
    \item \label{cc:ood_label_ood}It is beneficial to label OOD data as extra classes different from ID classes. If we can further label semantically different OOD data as different classes, the performance can be further improved.
\end{enumerate}

\section{Method}

Based on the findings in Section~\ref{sec:analysis}, we proposed $\Upsilon$-Model (named after its shape)  for class-mismatched SSL.
$\Upsilon$-Model trains a classifier $f$ that will output the posterior distribution over $K_{ID}+K$ classes,~\ie, $f(\y|\x)\in \mathbb{R}^{K_{ID}+K}, \1^\top f(\y|\x) = 1$. $K$ is the number of extra classes, which can be known in advance (\ie, $K=K_{OOD}$) or be set as a \textit{hyper-parameter}. Similar to vanilla PL, we define confidence with the same form as Equation~\ref{eq:confidence}. However, this confidence is a little different from its original definition in~\citet{Hendrycks17Baseline}, because we only calculate the maximum probability of the $K_{ID}$ classes instead of all. Therefore, we rename it to \textbf{In-Distribution confidence (ID confidence)}. For evaluation, we predict labels using Equation~\ref{eq:pred}. $\Upsilon$-Model aims to solve the following questions:
\begin{enumerate}[leftmargin=20mm,label=Problem \arabic*: ,ref=\arabic*,itemsep=0mm]
    \vspace{-1mm}
    \item \label{q:imblance}how to avoid imbalanced pseudo-labels in PL model? (Conclusion~\ref{cc:ce_imbalance},~\ref{cc:pl_imbalance})
    \item\label{q:id} how to avoid labeling OOD data as ID? (Conclusion~\ref{cc:ood_label_id})
    \item\label{q:ood} how to create proper pseudo-labels for unlabeled OOD data? (Conclusion~\ref{cc:ood_label_ood})
    \vspace{-1mm}
\end{enumerate}

\begin{figure}[t]
    \centering
    \includegraphics[width=\linewidth]{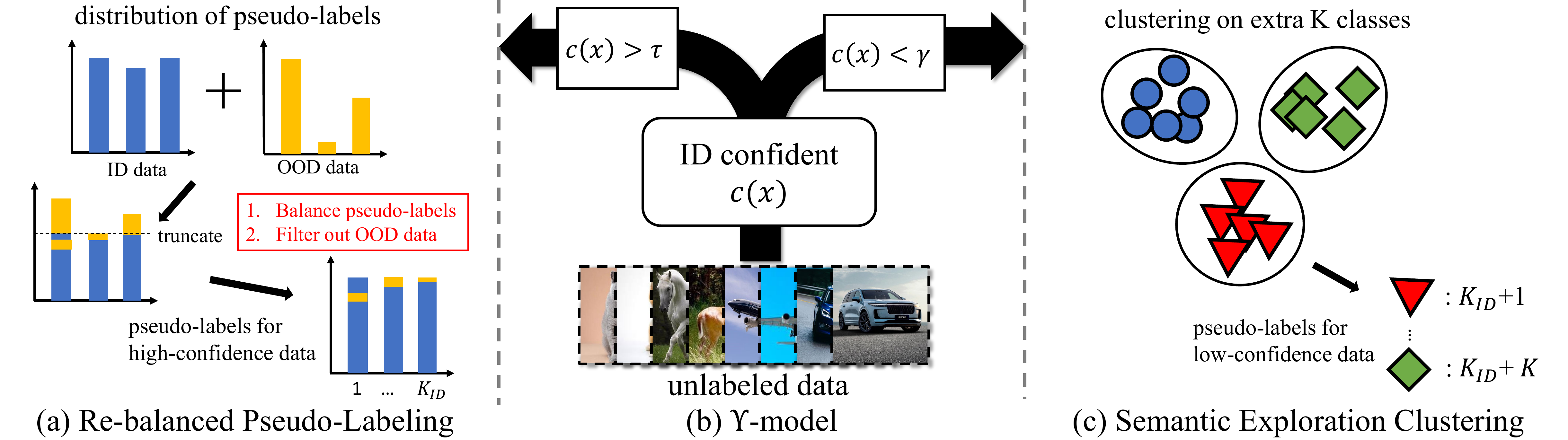}
    \vspace{-4mm}
    \caption{Illustration of $\Upsilon$-Model and its two main branches. (b) is the main structure of $\Upsilon$-Model where we judge by the ID confidence if certain unlabeled data belongs to ID classes or not. The data with high confidence will perform Re-balanced Pseudo-Labeling, while those with low confidence will get their pseudo-labels by Semantic Exploration Clustering. (a) Re-balanced Pseudo-Labeling truncates the number of pseudo-labeled data to the minimum, making the pseudo-labels balanced and filtering out OOD data. (c) Semantic Exploration Clustering simulates the process of learning from ground truth labels of OOD data, creating pseudo-labels on extra $K$ classes by clustering.}
    \vspace{-2mm}
    \label{fig:method}
\end{figure}

$\Upsilon$-Model consists of two main branches -- Re-balanced Pseudo-Labeling (RPL) and Semantic Exploration Clustering (SEC). RPL acts on high-confidence data to solve Problem~\ref{q:imblance},~\ref{q:id}. SEC acts on low-confidence data to solve Problem~\ref{q:ood}. We describe the two branches in the following sections. The overview of $\Upsilon$-Model is illustrated in Figure~\ref{fig:method}.

\vspace{-2mm}
\subsection{Re-balanced Pseudo-Labeling}
\label{subsec:rebalance}
\vspace{-2mm}
As illustrated in Section~\ref{subsec:pl_ood}, the main problem of vanilla PL is that a large number of OOD data with high confidence scores have imbalanced pseudo-labels. 
One possible solution is re-weighting the unlabeled sample~\citep{Guo20Safe} or using other methods in the imbalance learning field. However, even if we solve the problem of imbalance learning, labeling OOD data as ID classes also may damage the performance (Conclusion~\ref{cc:ood_label_id}). 
In this paper, 
we use a simple method -- Re-balanced Pseudo Labeling -- to simultaneously solve imbalance (Problem~\ref{q:imblance}) and incorrect recognition (Problem~\ref{q:id}). It produces a set $\mathcal{P}$ of pseudo-labeled samples in three steps:
\begin{align}
    &N = \min_{y \in \Y_{ID}} \left| \{\x \in \mathcal{D}_u \mid f(y\mid\x) > \tau \} \right| , & \label{eq:rpl_n} \\
    &\tau_y = \operatorname{select\_N-th}(\{f(y\mid \x)\mid \x \in \mathcal{D}_u \}) , 
    \label{eq:rpl_tau_y} \\
    &\mathcal{P} = \bigcup_{y \in \Y_{ID}} \{(\x,y) \mid f(y\mid\x) \ge \tau_y ,\x \in \mathcal{D}_u\},\label{eq:rpl_P}&
\end{align}
where $\operatorname{select\_N-th}$ returns the N-th biggest value of the given set. RPL first calculates the minimum number of pseudo-labeled samples for each ID class by Equation~\ref{eq:rpl_n}. Then it truncates the number of pseudo-labels of each ID class to that number by Equation~\ref{eq:rpl_tau_y},~\ref{eq:rpl_P}. The process of RPL is illustrated in Figure~\ref{fig:method}(a). First, it enforces the pseudo labels on ID classes to be balanced, solving Problem~\ref{q:imblance}. Second,  as is shown in Section~\ref{sec:analysis}, the set of high-confidence data is a mixture of ID and ODD data. Due to Conclusion~\ref{cc:ce_imbalance}, the pseudo-label distribution of such a set is a sum of imbalanced and balanced ones, thus still imbalanced. However, by selecting only top-$N$ confident samples for each ID class, we will keep ID data and omit many OOD data since confidence on ID data tends to be higher than OOD data~\citep{Hendrycks17Baseline}. This process solves Problem~\ref{q:id}. 

\vspace{-2mm}
\subsection{Semantic Exploration Clustering}
\label{subsec:clustering}
\vspace{-2mm}
As is demonstrated in Section~\ref{subsec:pl_ood}, if we know a set of samples is OOD, it will improve the performance if we label them as a unified class $K_{ID}+1$. But the best way is to use their ground truths (Conclusion~\ref{cc:ood_label_ood}). However, it is impossible to access their ground truths since they are unlabeled. We resort to using Deep Clustering methods~\citep{Caron18Deep,Asano20Self} to mine their semantics and approximate the process of learning with the ground truths. Modern Deep Clustering can learn semantically meaningful clusters and achieves competitive results against supervised learning~\citep{Gansbeke20Wouter}. 
Here, we use the balanced clustering method in~\citet{Asano20Self,Caron20Unsupervised} to create pseudo-labels for these OOD data. Assuming there are $M$ samples recognized as OOD, we first compute their soft targets:
\begin{equation}
    \begin{aligned}
        &\min _{Q \in U(K, M)}\langle Q,-\log P\rangle, \\ &U(K, M):=\left\{Q \in \mathbb{R}_{+}^{K \times M} \mid Q \1=\frac{1}{K}\1, Q^{\top} \1=\frac{1}{M}\1\right\} ,
        \label{eq:sela}
    \end{aligned}
\end{equation}
where $P\in \mathbb{R}_{+}^{K \times M},P_{ij} = \hat{f}(K_{ID}+i | \x_j)$. $\hat{f}$ is the normalized posterior distribution on extra classes,~\ie, $\hat{f}(K_{ID}+i | \x_j) = f(K_{ID}+i | \x_j) / \sum_{k=1}^{K} f(K_{ID}+k | \x_j)$.
We use \textit{the Sinkhorn-Knopp algorithm}~\citep{Cuturi13Sinkhorn} to optimize $Q$. Once we get $Q$, we harden the label by picking the class with the maximum predicted probability and mapping it to the extra $K$ classes:
\begin{equation}
    \hat{y}_j = K_{ID} + \argmax_{i} Q_{ij}.
    \label{eq:ood_hard_label}
\end{equation}
$\hat{y}_j$ is used as the pseudo-label for $\x_j$. We perform SEC on the set of data with ID confidence lower than a threshold $\gamma$,~\ie, $\{\x | c(\x) < \gamma\}$. It may be the concern that introducing a clustering component makes the $\Upsilon$-Model too computationally expensive to be practical. We give analyses of time complexity in ~\cref{app:time_complexity}.

\section{Related Work}
\vspace{-2mm}
\paragraph{Class-Mismatched Semi-Supervised Learning.} Deep Semi-Supervised Learning suffers from performance degradation when there are unseen classes in unlabeled data~\citep{Oliver18Realistic}. As the proportion of such out-of-distribution (OOD) data get larger, the performance drop more. To cope with such a class-mismatched problem, several methods are proposed. ~\citet{Chen20Semi} formulate a sequence of ensemble models aggregated accumulatively on-the-fly for joint self-distillation and OOD
filtering. ~\citet{Guo20Safe} re-weight the unlabeled data by meta-learning to decrease the negative effect of OOD data. ~\citet{Huang20They} recycle transferable OOD data employing adversarial learning. 
Recently,\citet{Saito21OpenMatch} proposed open-set consistency regularization to  improve outlier detection. ~\citet{Cao21Open} proposed open-world semi-supervised learning, where the classes of unlabeled data need to be discovered. From a methodological perspective, both \citet{Cao21Open} and our method use cluster to mining semantics. However, their method is originated from the demand of discovering novel classes, while ours is based on analyses of best labeling strategies. ORCA does not give such important analyses. Additionally, we solve the imbalanced pseudo-labels by RPL while they do not reveal this problem.
Different from all these methods, we conduct a comprehensive study on Pseudo-Labeling (PL) and give useful guidance on how to do better in class-mismatched SSL. In addition to that, we reveal the imbalance phenomenon and propose RPL. 
\vspace{-3mm}
\paragraph{Pseudo-Labeling.} The method of Pseudo-Labeling, also known
as self-training, is a simple and effective way for Deep SSL~\citep{Lee2013Pseudo,Shi18Transductive,Arazo2020Pseudo,Iscen19Label}. Despite its simplicity, it has been widely applied to diverse fields such as image classification~\citep{Xie20Self}, natural language processing~\citep{He20Revisiting} and object detection~\citep{Rosenberg05Semi}. The use of a hard label makes Pseudo-Labeling closely related to entropy minimization~\citep{Grandvalet04Semi}.
\vspace{-3mm}
\paragraph{Deep Clustering and Novel Class Discovery.} Deep clustering methods improve the ability of traditional cluster methods by leveraging the representation power of DNNs. A common means is to transform data into low-dimensional feature vectors and apply traditional clustering methods~\citep{Yang17Towards,Caron18Deep}. In Self-Supervised Learning, clustering methods are used to learn meaningful representation for downstream tasks~\citep{Caron18Deep,Asano20Self,Caron20Unsupervised}. Modern Deep Clustering can learn semantically meaningful clusters and achieves competitive results against supervised learning~\citep{Gansbeke20Wouter}. With the help of Deep Clustering, the concept of Novel Class Discovery (NCD) was first formally introduced by~\citet{Han19Learning}. 
There are two main differences between NCD and our setting. First, the goal of NCD is to correctly cluster OOD data while semi-supervised learning aims to correctly classify ID data. Second, NCD knows which data are OOD in advance while class-mismatched SSL does not. Therefore, \textit{NCD methods can not directly apply to this setting}. A recent method, UNO [1], uses a similar cluster method to discover semantics among OOD data like our SEC component. However, since they mainly focus on cluster OOD data, they do not reveal the benefit for classifying ID data when mining the semantics of OOD data, which is one of the contributions of this paper.
\section{Experiments}
\label{sec:exp}
\paragraph{Dataset.} We test our methods on the five datasets as in~\cref{sec:analysis}, \ie, \textbf{CIFAR10 (6/4)}, \textbf{SVHN (6/4)}, \textbf{CIFAR100 (50/50)}, \textbf{Tiny ImageNet (100/100)}
and \textbf{ImageNet100 (50/50)}.
The class-mismatched ratio is set as $\{0\%,25\%,50\%,75\%,100\%\}$.
\vspace{-3mm}
\paragraph{Implementation Details.} We use the same network and training protocol as~\cref{sec:analysis}. We first train a classification model only on labeled data for 100 epochs without RPL and SEC. We update pseudo-labels every 2 epochs. For both datasets, we set $\tau=0.95$, $\gamma=0.3$,$K=4$. We use an exponential moving average model for final evaluation as in~\citet{Athiwaratkun19There}.

\begin{figure}[htp]
	\centering
	\setcounter{subfigure}{0}
	\subfigure[]{
		\label{subfig:ssl_cifar}
		\includegraphics[width=0.22\linewidth]{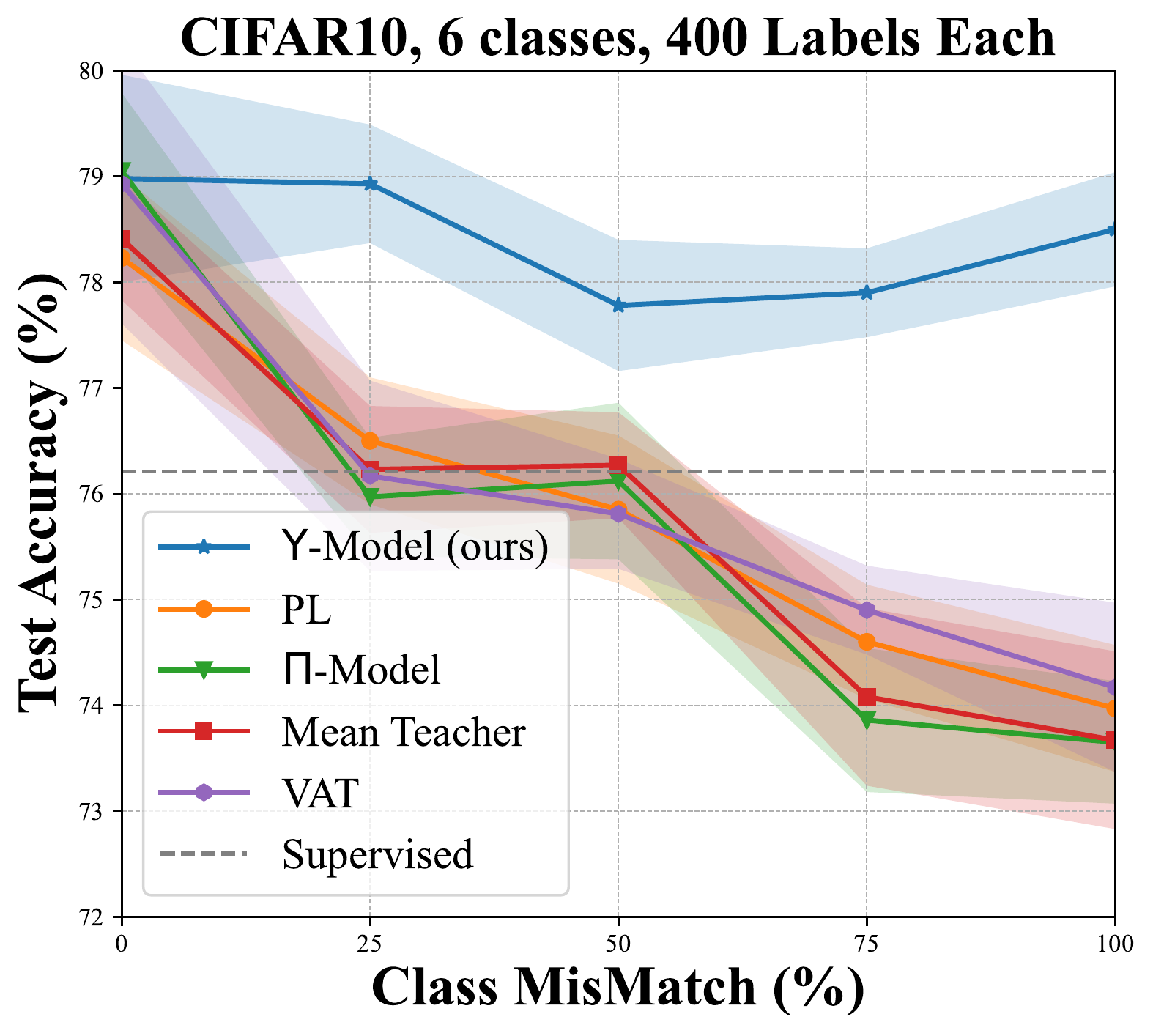}
	}
	\subfigure[]{
		\label{subfig:ssl_svhn}
		\includegraphics[width=0.22\linewidth]{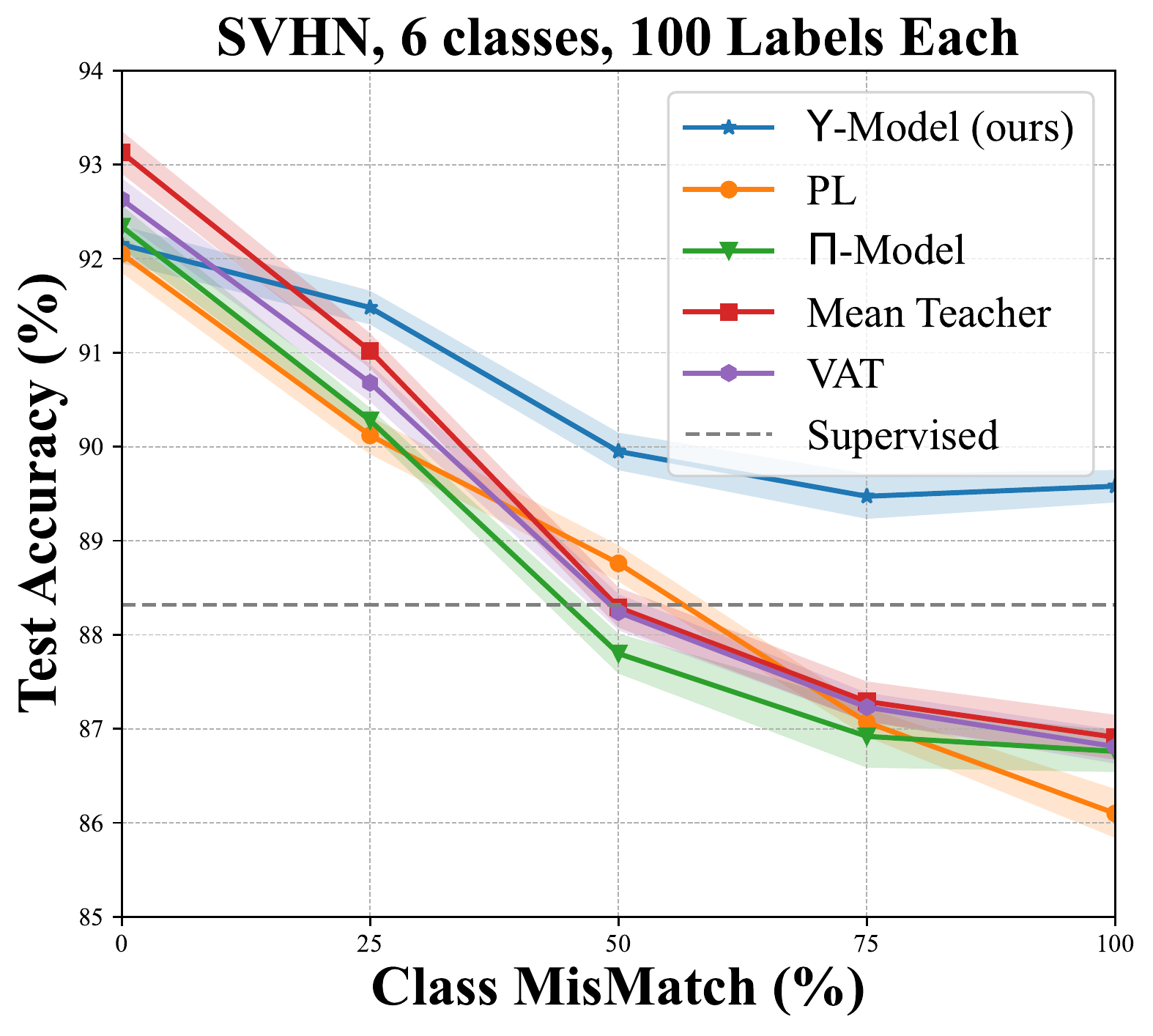}
	}
	\subfigure[]{
		\label{subfig:cmssl_cifar}
		\includegraphics[width=0.22\linewidth]{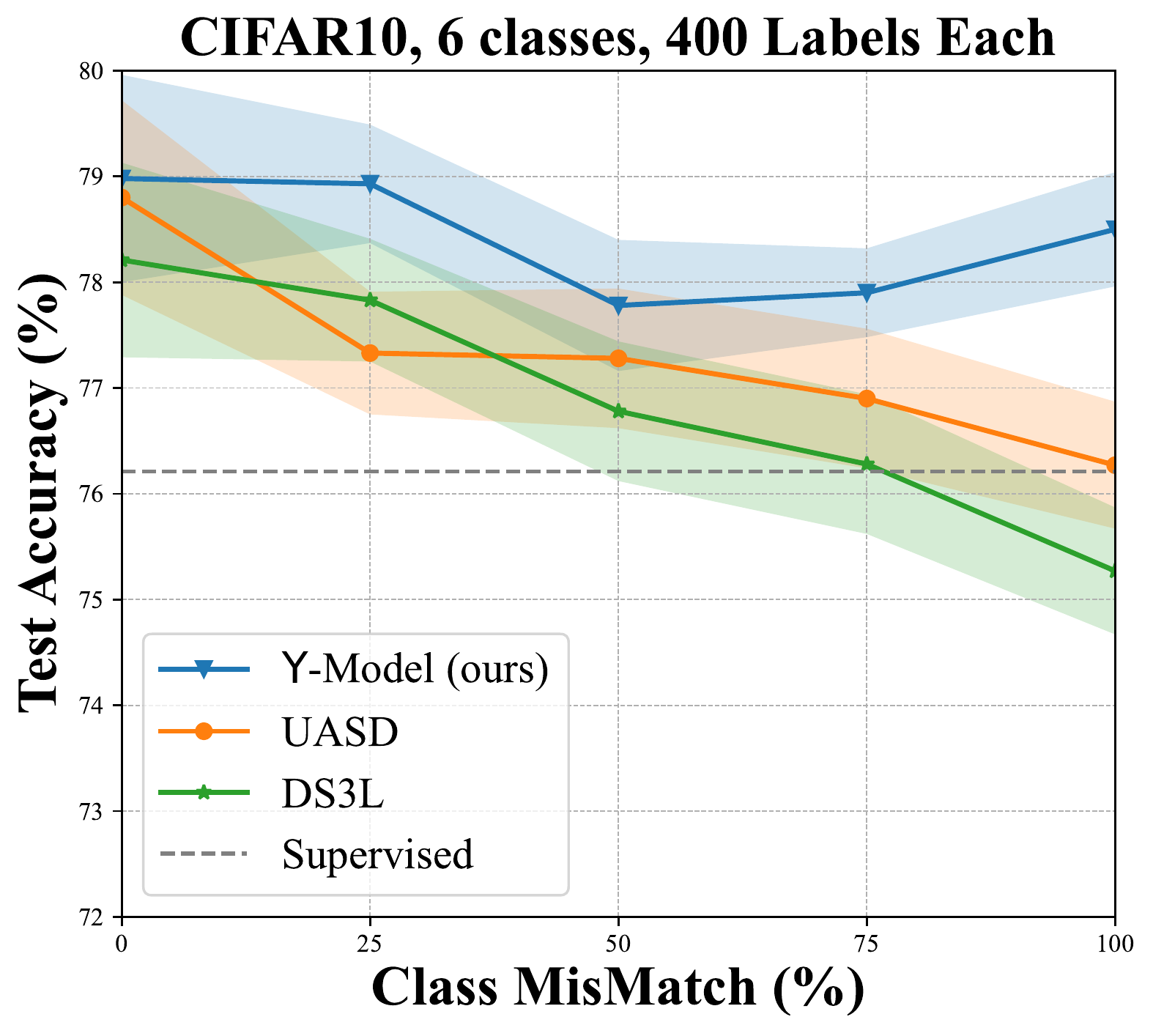}
	}
	\subfigure[]{
		\label{subfig:cmssl_svhn}
		\includegraphics[width=0.22\linewidth]{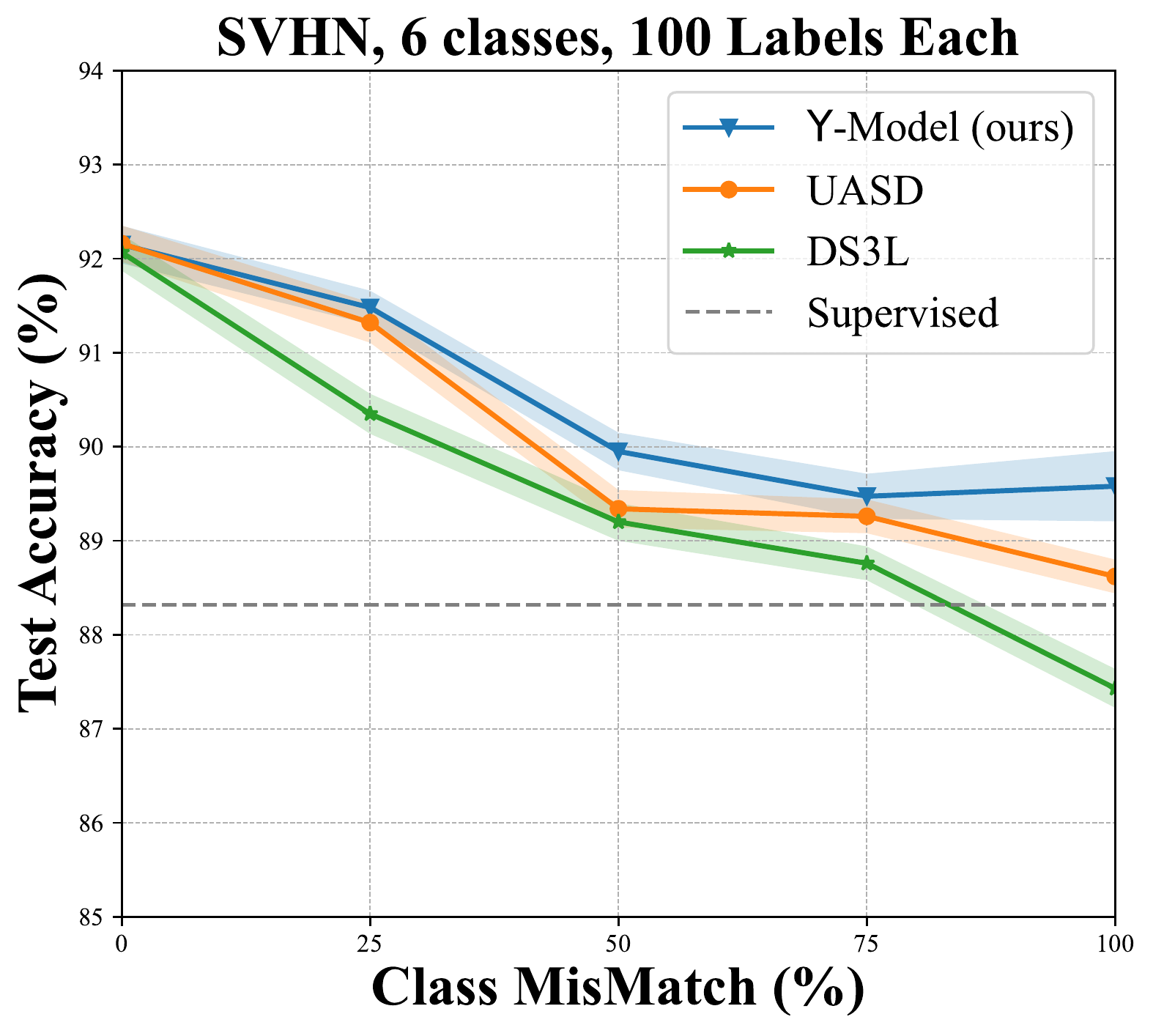}
	}
	\vspace{-2mm}
	\caption{Comparison with existing methods on CIFAR10 and SVHN dataset with Wide-ResNet-28-2 network. Class mismatch ratios are varied. The shaded regions with the curves indicate the standard deviations of the accuracies over five runs. (a) (b) Comparison with traditional SSL methods. These methods suffer from performance degradation as the mismatch ratio increases. (c) (d) Comparison to two existing class-mismatched SSL methods -- UASD and DS$^3$L. Our methods perform better in almost all the experimental setups.}
	\vspace{-2mm}
\end{figure}

\begin{table}[h]
\label{tb:compare_in}
	\caption{Comparison of three mid/large-scale datasets with different class-mismatched ratios. The backbone is Wide-ResNet-28-2 for all experiments. The standard deviations of the accuracies over five runs are also reported. The best results are highlighted in \textbf{bold}. The second-best results are highlighted in \underline{underline}. 
	Baseline and Oracle have the same meaning as in~\cref{subsec:pl_ood}.}
	\centering
	\begin{threeparttable}
		\begin{tabular}{cccccc}
			\toprule
			Mismatch Ratio       & $0\%$        & $25\%$       & $50\%$       & $75\%$       & $100\%$      \\
			\midrule
			& \multicolumn{5}{c}{\textbf{CIFAR100 (50/50)}}                                   \\
			\midrule
			PL       & 61.68 $\pm$ 0.30 & 60.20 $\pm$ 0.25 & 60.12 $\pm$ 0.30 & 57.79 $\pm$ 0.27 & 57.62 $\pm$ 0.57 \\
			UASD     & 60.30 $\pm$ 0.26 & 59.42 $\pm$ 0.61 & 59.92 $\pm$ 0.53 & 58.94 $\pm$ 0.63 & 58.74 $\pm$ 0.50 \\
			DS$^3$L     & 60.68 $\pm$ 0.67 & 60.60 $\pm$ 0.40 & 59.22 $\pm$ 0.33 & 59.74 $\pm$ 0.37 & \textbf{59.56 $\pm$ 0.57} \\
			$\Upsilon$-Model    & \textbf{62.10 $\pm$ 0.30} & \textbf{61.26 $\pm$ 0.40} & \underline{60.68 $\pm$ 0.24} & \underline{60.12 $\pm$ 0.38} &\underline{ 59.46 $\pm$ 0.36} \\
		 	ORCA &58.94 $\pm$ 0.24 &  59.98  $\pm$ 0.35 &  60.14 $\pm$ 0.40 & 58.84 $\pm$ 0.27 & 44.00 $\pm$ 0.42 \\
            OpenMatch & \underline{62.08 $\pm$ 0.26} & \underline{60.94 $\pm$ 0.36} & \textbf{60.92 $\pm$ 0.23} & \textbf{60.36 $\pm$ 0.47} & 59.22 $\pm$ 0.50\\
			Baseline & \multicolumn{5}{c}{58.68 $\pm$ 0.25}                                 \\
			Oracle &75.46 $\pm$ 0.20&73.36 $\pm$ 0.29 & 72.94 $\pm$ 0.25 & 69.26 $\pm$ 0.26 & 63.90 $\pm$ 0.30 \\
			\midrule
			& \multicolumn{5}{c}{\textbf{Tiny ImageNet (100/100)}}                              \\
			\midrule
			PL       & 43.42 $\pm$ 1.03 & \underline{42.88 $\pm$ 1.51} & 41.94 $\pm$ 1.49 & 39.72 $\pm$ 2.30 & 38.94 $\pm$ 2.41 \\
			UASD     & 43.34 $\pm$ 0.78 & 42.34 $\pm$ 0.61 & 41.80 $\pm$ 1.33 & 41.08 $\pm$ 1.16 & 36.16 $\pm$ 1.05 \\
			DS$^3$L$^\ast$     &      -      &       -     &       -     &       -     &        -    \\
			$\Upsilon$-Model    & \textbf{44.42 $\pm$ 0.43} & \textbf{43.48 $\pm$ 0.40} & \textbf{42.42 $\pm$ 0.95} & \textbf{43.22 $\pm$ 0.60} & \textbf{41.76 $\pm$ 0.63} \\
			ORCA & 42.16 $\pm$ 0.41 & 40.22 $\pm$ 0.37 & 41.58 $\pm$ 1.24 &\underline{42.92 $\pm$ 0.57} & 40.82 $\pm$ 0.36 \\
			OpenMatch       & \underline{43.42 $\pm$ 0.31} & 42.00 $\pm$ 0.37 & \underline{42.26 $\pm$ 0.27} & 42.5 $\pm$ 0.31 & \underline{41.03 $\pm$ 0.38} \\
			Baseline & \multicolumn{5}{c}{39.66 $\pm$ 0.52}    \\
			Oracle &56.18 $\pm$ 0.31 & 53.8  $\pm$ 0.28 & 51.82 $\pm$ 0.20 &48.3 $\pm$ 0.30 & 45.28 $\pm$ 0.31 \\
			\midrule
			& \multicolumn{5}{c}{\textbf{ImageNet (50/50)}}  \\
			\midrule
			PL       & 50.04 $\pm$ 0.11 & 49.32 $\pm$ 0.15 & 48.36 $\pm$ 0.19 & 47.40 $\pm$ 0.13 & 46.96 $\pm$ 0.19 \\
			UASD     & \underline{50.24 $\pm$ 0.20} & 48.68 $\pm$ 0.26 & 48.12 $\pm$ 0.17 & 48.00 $\pm$ 0.11  & 47.04 $\pm$ 0.25 \\
			DS3L     &      -      &        -    &      -      & -           &                   -               \\
			$\Upsilon$-Model    & \textbf{50.6 $\pm$ 0.19}  & \underline{49.78 $\pm$ 0.16} & \textbf{48.88 $\pm$ 0.30} & \textbf{48.36 $\pm$ 0.11} &\textbf{47.64 $\pm$ 0.22}                       \\
			ORCA &41.56 $\pm$ 0.27 & 43.32 $\pm$ 0.38 & 46.60  $\pm$ 0.23 & 47.08 $\pm$ 0.45 & \underline{46.88 $\pm$ 0.36} \\
            OpenMatch & 49.92 $\pm$ 0.11 & \textbf{49.88 $\pm$ 0.15} &\underline{48.82 $\pm$ 0.19} & \underline{48.01 $\pm$ 0.21} & 46.84 $\pm$ 0.27      \\
			Baseline & \multicolumn{5}{c}{48.12 $\pm$ 0.29}   \\
			Oracle & 62.92 $\pm$ 0.26 &62.00 $\pm$ 0.28&61.04 $\pm$ 0.26 &57.92 $\pm$ 0.26 & 55.96 $\pm$ 0.30 \\
			\bottomrule
		\end{tabular}
		\begin{tablenotes}
			\footnotesize
			\item[$\ast$] We cannot finish DS$^3$L on mid-scale or large-scale datasets like Tiny ImageNet or ImageNet100 within a reasonable time. 
		\end{tablenotes}
	\end{threeparttable}
	\vspace{-4mm}
\end{table}

\subsection{Compare with Traditional SSL methods}
In this subsection, we compare our methods with four traditional SSL methods -- Pseudo-Labeling~\citep{Lee2013Pseudo}, $\Pi$-Model~\citep{Laine17Temporal}, Mean Teacher~\citep{Tarvainen17Mean} and VAT~\citep{Miyato19Virtual}. \footnote{We note that there are also other methods like FixMatch~\citep{Sohn20Fix} and MixMatch~\citep{Berthelot19Mix} focus on augmentation or other tricks. Their contribution is orthogonal to ours. Also, FixMatch is unstable in this setting. We provide comprehensive comparisons with these methods in~\cref{sec:fixmix}.
}~\cref{subfig:ssl_cifar},~\ref{subfig:ssl_svhn} show the results. Traditional methods suffer from performance degradation as the mismatch ratio increases. They usually get worse than the supervised baseline when the mismatch ratio is larger than $50\%$ on CIFAR10 and SVHN. In contrast, our methods get steady improvement under all class mismatch ratios. The reasons can be attributed as follows. First, our method is aware of the existence of OOD data. We do not treat OOD data like ID data, which can hurt performance. Second, we reuse OOD data by exploring their semantics which proves to be useful in ~\cref{subsec:pl_ood}. Therefore, even when the class-mismatched ratio gets $100\%$, the performance of $\Upsilon$-Model is still better than the supervised baseline.

\subsection{Compare with Class-Mismatched SSL methods}
\label{subsec:exp_cm}
In this subsection, we compare our method with two existing class-mismatched SSL methods -- UASD~\citep{Chen20Semi} and DS$^3$L~\citep{Guo20Safe}. For a fair comparison, we use Pseudo-Labeling as the base method of DS$^3$L. From ~\cref{subfig:cmssl_cifar},~\cref{subfig:cmssl_svhn}, we can see our methods are superior to these two methods in all settings. It is noticeable that DS$^3$L underperforms supervised baseline when all the unlabeled data are drawn from OOD classes. This is attributed to the fact that DS$^3$L uses a down weighting strategy to alleviate the negative effect of OOD data and does not change the form of unsupervised loss. But we have shown in \cref{subsec:pl_ood} that labeling OOD data as ID classes damages performance anyhow. On the contrary, $\Upsilon$-Model uses the OOD data in the right way -- simulating the process of training them with their ground truth labels. As a result, our method shows superiority especially under a large class-mismatched ratio. We also notice that the performance curve of $\Upsilon$-Model appears a U-shape (obvious on CIFAR10). A possible reason is that RPL and SEC compete with each other. RPL tends to make samples get a high prediction on ID classes while SEC tends to make samples get a high prediction on OOD classes. When the class-mismatched ratio reaches $0\%$ ($100\%$), RPL (SEC) dominates the other. In this circumstance, one works without any disturbance to the other. However, when the class-mismatched ratio is $50\%$, they compete fiercely with each other, causing many incorrectly recognized ID or OOD samples. 

Additionally, we compare our method to OpenMatch~\citep{Saito21OpenMatch} and ORCA~\citep{Cao21Open} on mid-/large-scale datasets in~\cref{tb:compare_in}. \textit{Again we use the same augmentation for both labeled and unlabeled data to avoid unfairness for baseline}. ORCA is an open-world semi-supervised learning method, which not only aims to correctly classify ID data but also to cluster OOD. Due to this, its ID classification performance is lower than others. OpenMatch is a competitive method, but it can not surpass ours in most of the settings.

\subsection{Ablation Study}
In this section, we validate the functionality of RPL and SEC. We conduct experiments on CIFAR10 benchmark as in the analysis section~\ref{sec:analysis}. 

\begin{table}[h]
	\setlength\tabcolsep{5pt}
	\centering
	\caption{Validation of RPL and SEC. The experiments are conducted on CIFAR10 with a Wide-ResNet-28-2 backbone. Class-mismatched ratio varies from 0 to $100\%$. Check RPL or not means using RPL or vanilla PL. $K=0$ means we do not use SEC. $K=1$ means we label all the low confidence data as a unified class $K_{ID}+1$.}
	\begin{tabular}{cccccccc}
		\toprule
		&                       &                     & \multicolumn{5}{c}{Class-Mismatched Ratio (\%)}                         \\ \cmidrule(lr){4-8}
		\multirow{-2}{*}{RPL} & \multirow{-2}{*}{SEC} & \multirow{-2}{*}{$K$} & 0    & 25   & 50                           & 75   & 100    \\
		\midrule 
		&     & 0 & 78.23 $\pm$ 0.39                                & 76.50 $\pm$ 0.30                 & 75.85 $\pm$ 0.35               & 74.60 $\pm$ 0.27                 & 73.97 $\pm$ 0.30               \\ 
		\checkmark&     & 0 & \textbf{78.76 $\pm$ 0.49}                                & 76.92 $\pm$ 0.32                & 76.80 $\pm$ 0.31                & 75.48 $\pm$ 0.27                & 75.38 $\pm$ 0.27               \\ 
		\checkmark&   \checkmark  & 1 & \textbf{78.86 $\pm$ 0.37}                                & 77.17 $\pm$ 0.55                & 77.12 $\pm$ 0.37               & 76.88 $\pm$ 0.35                & 77.27 $\pm$ 0.39               \\ 
		&   \checkmark  & 4 & 78.11 $\pm$ 0.45                                & 77.43 $\pm$ 0.29                & 77.46 $\pm$ 0.37               & 76.38 $\pm$ 0.37                & 74.18 $\pm$ 0.32               \\ 
		\checkmark&  \checkmark   & 4 & \textbf{78.98 $\pm$ 0.49}                               &\textbf{78.93 $\pm$ 0.28}                & \textbf{77.78 $\pm$ 0.31}               & \textbf{77.90 $\pm$ 0.21}                 & \textbf{78.50 $\pm$ 0.27}                \\ 
		\midrule
		\multicolumn{3}{c}{Baseline}  & \multicolumn{5}{c}{76.21 $\pm$ 0.21} \\                                                                                                                                               
		\bottomrule
	\end{tabular}
	\vspace{-2mm}
	\label{tb:ablation}
\end{table}
\vspace{-2mm}
\paragraph{Validation of effectiveness of RPL and SEC.}  We conduct ablation studies under different class-mismatched ratios and report the averaged test accuracy and standard deviation of five runs. As usual, we vary the class-mismatched ratio. Table~\ref{tb:ablation} displays the results. Firstly, comparing the first line and second line of the table, RPL not only outperforms vanilla PL in high class-mismatched ratio scenarios but also improves in low class-mismatched ratio scenarios. This reveals that balanced pseudo-labels always help since once the model creates imbalanced pseudo-labels, it will deteriorate when there are not enough measures to correct it. Secondly, comparing the second and third line, it shows that RPL alone alleviate the performance degradation but it can not prevent it, in accord with Conclusion~\ref{cc:ood_label_id}. When using SEC, $\Upsilon$-Model gets better results than supervised baseline when the class-mismatched ratio is high. Besides, comparing the third and last lines, we see that when clustering OOD data and exploring their semantics instead of using a unified class to label them, the performance improves.

\begin{figure}[tp]
	\centering
	\setcounter{subfigure}{0}
	\subfigure[]{
		\label{subfig:pl_ratio_line}
		\includegraphics[width=0.31\linewidth]{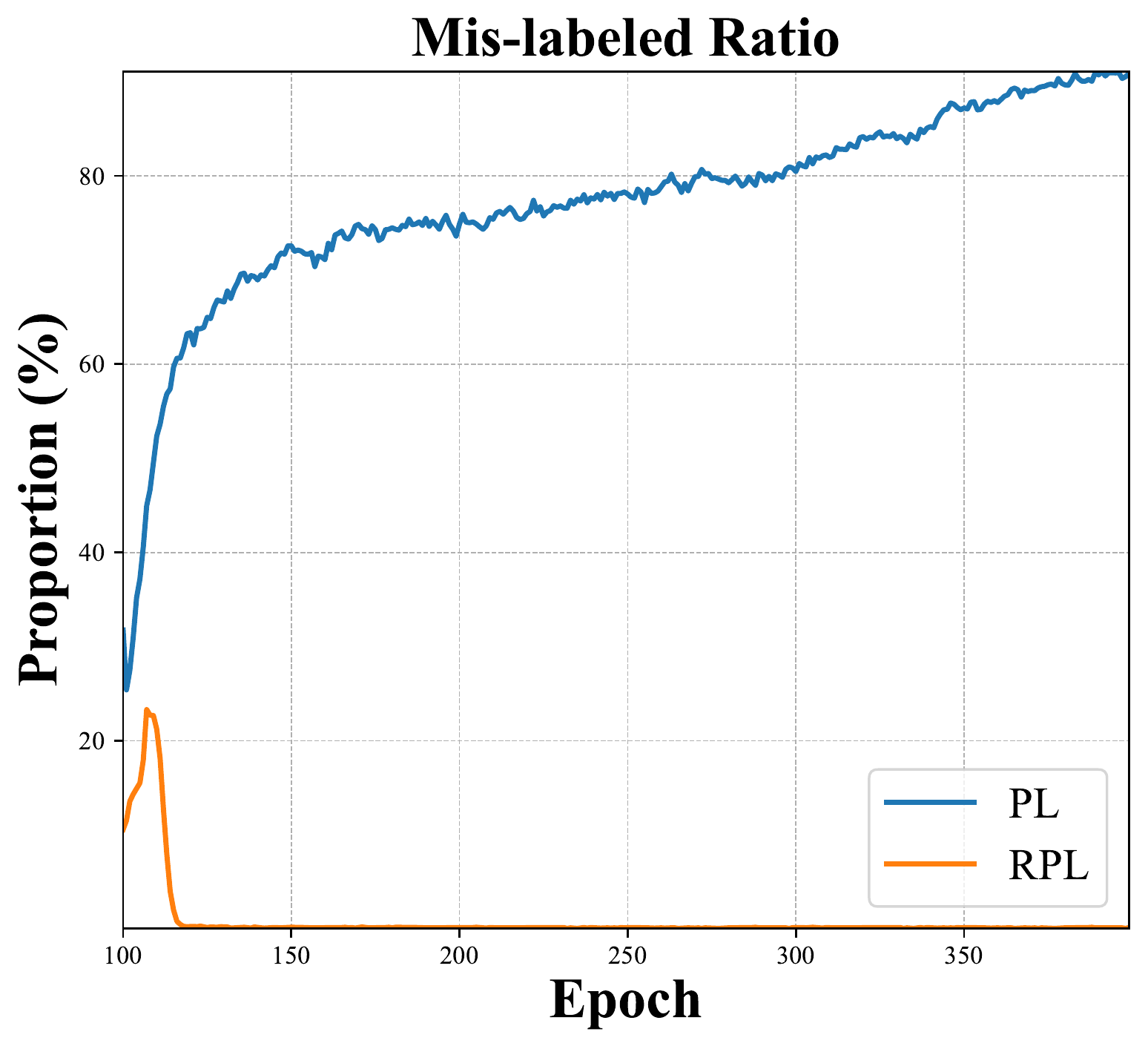}
	}
	\hspace{-4mm}
	\subfigure[]{
		\label{subfig:cm_model}
		\includegraphics[width=0.31\linewidth]{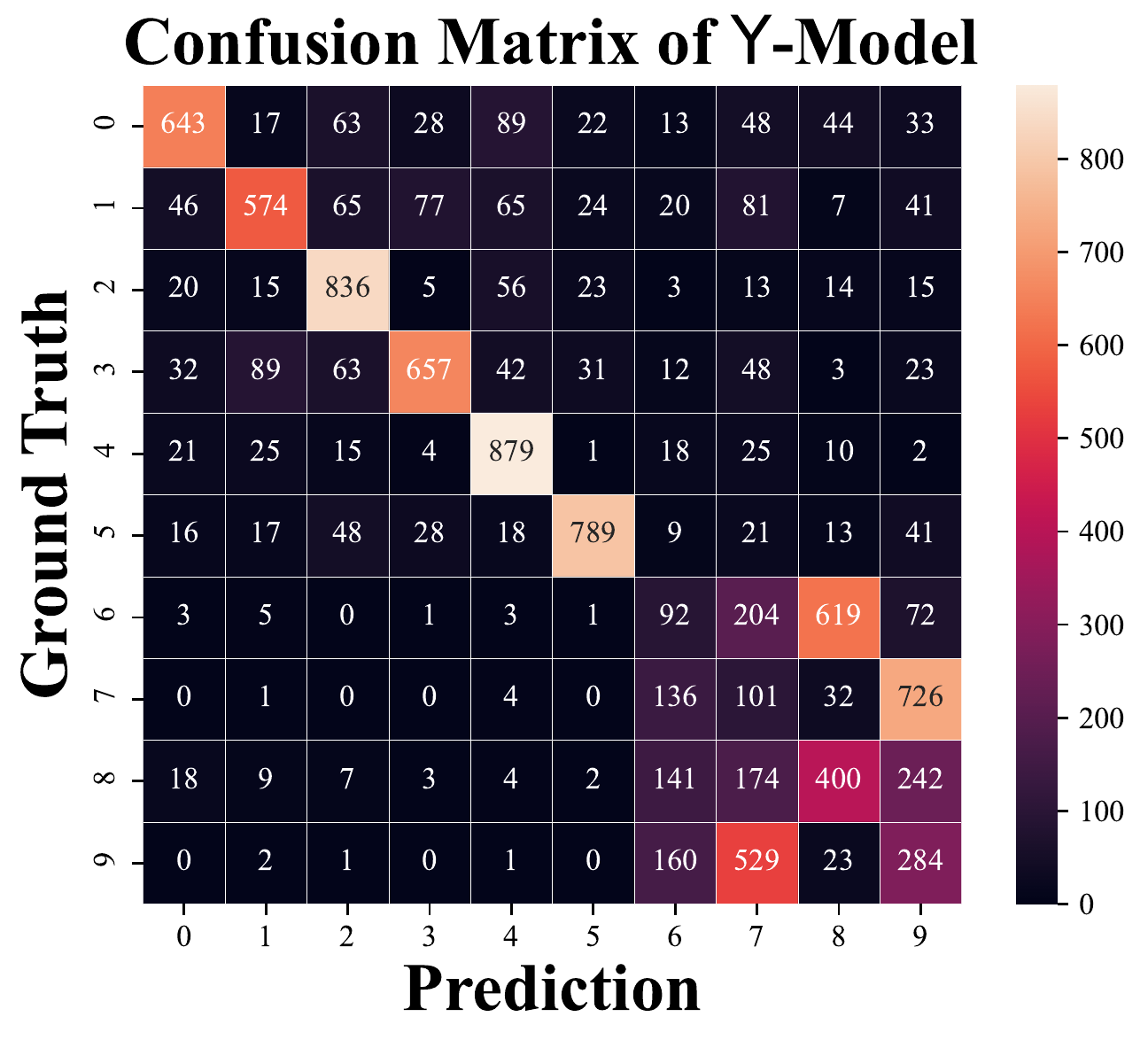}
	}
	\subfigure[]{
		\label{subfig:K_line}
		\includegraphics[width=0.31\linewidth]{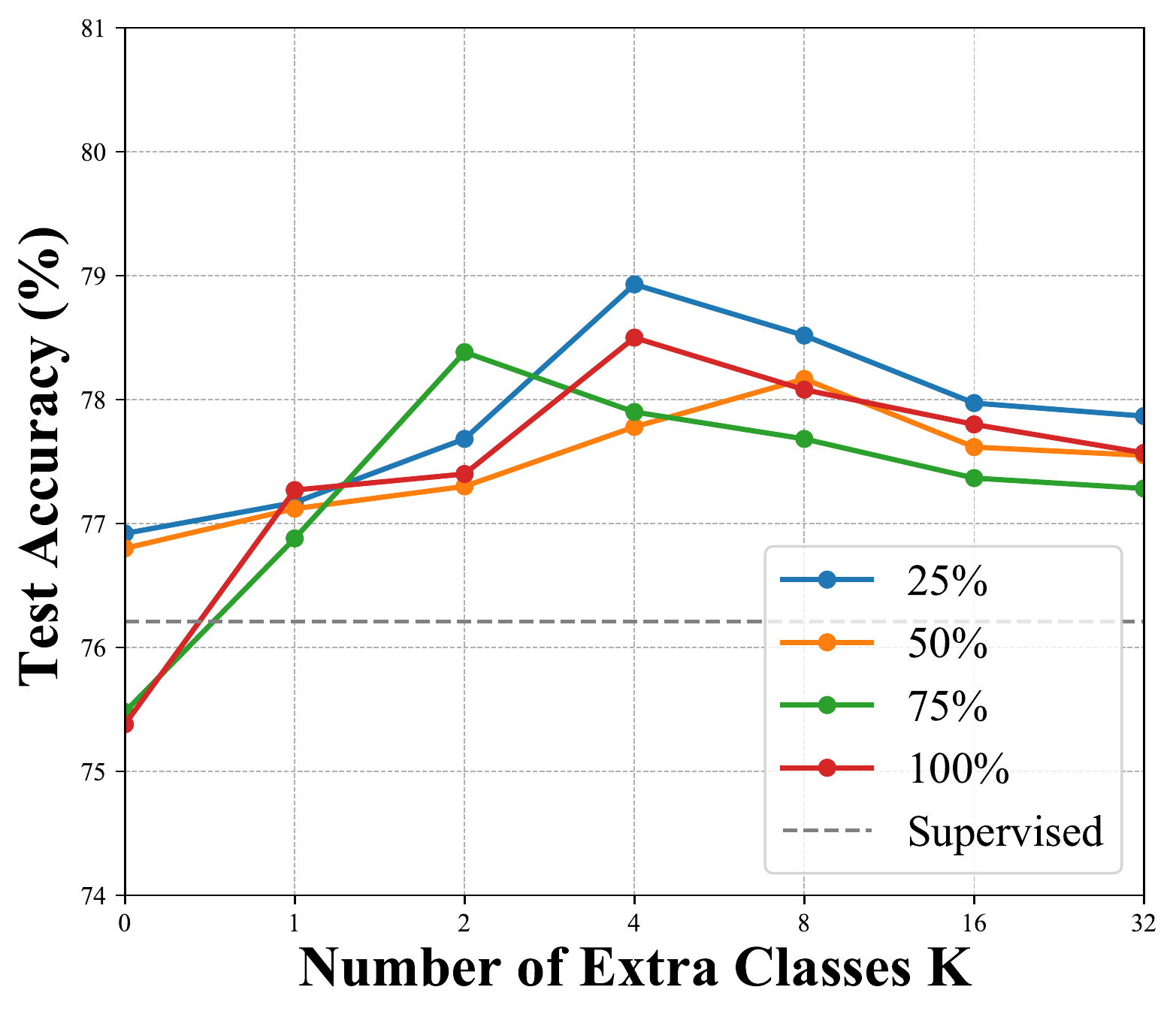}
	}
	\vspace{-4mm}
	\caption{Two experiments on CIFAR10 benchmark to validate the functionality of RPL and SEC. (a) The proportion of OOD data pseudo-labeled as ID classes with the training goes on. Vanilla PL makes this ratio keep increasing while RPL makes it drop to 0. It demonstrates that RPL help filter out OOD data. (b) Confusion matrix on the full test set of CIFAR10. RPL solves the imbalance problem.}
	\vspace{-3mm}
\end{figure}
\vspace{-2mm}
\paragraph{RPL helps filter out OOD data and solve the imbalance problem.} It is noticeable that the last two lines of Table~\ref{tb:ablation} show that without RPL, SEC alone can not achieve better performance than supervised baseline. We show the reason here. Figure~\ref{subfig:pl_ratio_line} plots the proportion of OOD data that are pseudo-labeled as ID classes. Without RPL,~\ie, using vanilla PL, the number of incorrectly recognized OOD data keep increasing as the training proceeds. While with RPL, this ratio rapidly drops to 0. This proves that RPL help filter out OOD data by utilizing the imbalance property of OOD data. Further, we present the confusion matrix of $\Upsilon$-Model on the full test set (all the 10 classes) of CIFAR10. Compared to vanilla PL in  Figure~\ref{fig:subfig:cm_pl}, $\Upsilon$-Model does not suffer from imbalance problem, as a result of which, its performance is not degraded.

\vspace{-2mm}
\paragraph{Effect of extra class number $K$.} We vary the number of extra classes $K$. \cref{subfig:K_line} shows the result on CIFAR10 with various class mismatch ratios. Without SEC ($K=0$), $\Upsilon$-Model underperforms the supervised baseline. Using SEC ($K\ge 1$), $\Upsilon$-Model is always better than baseline. Also, it reaches its best performance when $K$ is roughly the actual number of OOD classes. This demonstrates that by simulating the process of training on OOD data with ground truth, SEC helps the classification model on ID data. Also, our model benefits from SEC but is not sensitive to the selection of $K$. SEC helps explore the semantic structure of OOD data, but it does not affect the learning on ID data since they perform tasks on different classifiers. 

\begin{table}[htbp]
	\centering
	\caption{Comparison of PL and $\Upsilon$-Model on the constructed class-mismatched CIFAR10 dataset with \underline{imbalanced OOD data}.}
	\begin{tabular}{ccccc}
		\toprule
		Class-Mismatched Ratio (\%) & 25 & 50 & 75 & 100  \\
		\midrule
		PL &  76.42 & 75.78 & 74.58 & 73.33 \\
		$\Upsilon$-Model (Ours) &\textbf{77.57}&\textbf{77.42}&\textbf{77.13}&\textbf{77.26} \\
		\bottomrule
	\end{tabular}
	\label{tb:im}
	\vspace{-2mm}
\end{table}
\vspace{-3mm}
\paragraph{Effectiveness on imbalanced OOD data.}
It might be a question that whether it is suitable to use balanced clustering since the OOD data are usually imbalanced in the real world. We note that we use cluster methods only to mine the semantics among OOD data. We do not aim to correctly recognize them. The effectiveness of using clustering methods to learn semantically meaningful representations has been proved in many works~\citep{Asano20Self,Caron20Unsupervised}. To prove the effectiveness of our method, we conduct experiments on an additional benchmark \textit{where the OOD data is imbalanced}. The OOD data are subsampled like CIFAR-10-LT~\citep{Cao19Learning} with imbalance ratio of $10$ on the OOD classes. \cref{tb:im} displays the results. We can see that our model can handle situations where OOD data are even imbalanced.  

\vspace{-3mm}
\paragraph{Other imbalance methods and uncertainty measures.} We present results of an imbalanced method applied to PL : PL-Reweight and two uncertainty measures applied to our model: $\Upsilon$-Model (Ent) and $\Upsilon$-Model (SD). Ent and SD detect OOD data via entropy and score difference. Ent uses negative entropy of the output distribution as the confident measure. SD use the difference between the largest  and the second-largest output probability as the confidence score.
\begin{table}[h]
	\centering
	\caption{Experiments on combining different imbalance and OOD detection methods with PL/$\Upsilon$-Model. PL-Reweight uses reweight strategy to balance pseudo-labels. Ent (Entropy) uses negative entropy of the output distribution as the confident measure. SD (Score Difference) uses the difference between the largest  and the second-largest output probability as the confidence score.}
	\begin{tabular}{ccccc}
		\toprule
		Class-Mismatch Ratio (\%) & 25 & 50 & 75 & 100  \\
		\midrule
		PL &  76.5&75.85&74.6&73.97 \\
		PL-Reweight &  76.98 & 76.33& 74.52 & diverge \\
		$\Upsilon$-Model (Ent) &  77.78 &77.97 &76.32 &77.12  \\
		$\Upsilon$-Model (SD)&  77.70 &\textbf{78.25} &\textbf{78.02} &76.72 \\
		$\Upsilon$-Model (Ours) &\textbf{78.93} &77.78&77.90 &\textbf{78.50 }\\
		\bottomrule
	\end{tabular}
\end{table}

Common imbalance methods like resample and reweight will be \textit{unstable} in this setting since there is the possibility that the number of pseudo-labels is zero in certain classes. Also, in~\cref{subsec:summary}, we have concluded that assigning ID labels for OOD data will degrade the model. Therefore, solving the imbalance problem is only a partial solution for this setting. In contrast, our RPL is stable in this situation and uses the imbalance phenomenon to simultaneously filter out OOD data. Also, uncertainty measures other than confidence score are applicable to our method.



\section{Conclusion}
\label{sec:conclusion}
In this paper, we analyze Pseudo-Labeling in class-mismatched semi-supervised learning where there are unlabeled OOD data from other classes. 
We show that Pseudo-Labeling suffers from performance degradation due to imbalanced pseudo-labels on OOD data. The correct way to use OOD data is to label them as classes different from ID classes while also partitioning them according to their semantics. 
Based on the analysis, we proposed $\Upsilon$-Model and empirically validate its effectiveness. 

In future work, we will explore whether other forms of semi-supervised learning methods like the consistency-based method suffer from the same problems. Also, currently our model is based on Pseudo-labeling. But we see the possibility of extending it to other SSL methods. For example, there may be re-balanced consistency methods and SEC may be a plug-and-play component for all SSL methods.
\section{Acknowledgments}

This work is partially supported by The National Key
R\&D Program of China (2020AAA0109401), NSFC (62006112, 61773198, 61921006), NSF of Jiangsu Province (BK20200313), CCF-Baidu Open Fund (NO.2021PP15002000).

\bibliography{references}
\bibliographystyle{tmlr}
\newpage

\appendix

\section{Algorithm}
\vspace{-4mm}
\begin{algorithm}[htp]
    \caption{$\Upsilon$-Model algorithm}
    \label{alg:model}
    \begin{algorithmic}[1]
        \Require
        Labeled dataset $\mathcal{D}_l = \{(\x_{li},y_{li})\}^n_{i=1}$, and unlabeled dataset $\mathcal{D}_u = \{\x_{ui}\}^m_{i=1}$; 
        Classification model $f_\phi$ parameterized with $\phi$, 
        ID class number $K_{ID}$, extra class number $K$, total epoch number $E$, pretrain epochs $E_{pt}$, interval to update pseudo-labels $E_{pl}$, pseudo-labeled set $\mathcal{P}$, confidence calculation function $c$.
        \Function{RebalancedPseudoLabeling}{$\mathcal{D},f,\tau$}
        \State $N  \gets \min_{y \in \Y_{ID}} \left| \{\x \in \mathcal{D}_u \mid f(y\mid\x) > \tau \} \right|$
        \State $\tau_y \gets \mathop{Nth\_biggest}(\{f(y\mid \x)\mid \x \in \mathcal{D}_u \}) , \quad  y = 1,2,\dots,K_{ID}$
        \State $\mathcal{P} \gets \bigcup_{y \in \Y_{ID}} \{(\x,y) \mid f(y\mid\x) \ge \tau_y ,\x \in \mathcal{D}_u\}$
        \State \Return $\mathcal{P}$
        \EndFunction
        \Function{SemanticExplorationClustering}{$\mathcal{D},f,\gamma$}
        \State $S \gets \{\x | c(x) < \gamma\}$
        \State $M \gets \lvert S \rvert$
        \State $P_{ij} \gets f(K_{ID}+i | \x_j) / \sum_{k=1}^{K} f(K_{ID}+k | \x_j), \quad i = 1,2,\dots,K, \quad j = 1,2,\dots,M $
        \State Solve~\ref{eq:sela} by Sinkhorn-Knopp algorithm and get $Q$
        \State $\hat{y}_j \gets K_{ID} + \argmax_{i} Q_{ij}, \quad j = 1,2,\dots,M$
        \State $\mathcal{C} \gets \{(\x_j,\hat{y}_j)\}_{j=1}^M$
        \State \Return $\mathcal{C}$
        \EndFunction
        \For{e = 1 to $E$}
        \If{e $< E_{pt}$}
        \State train $f_\phi$ with standard supervised learning on $\mathcal{D}_l$ \Comment{Pre-training phase}
        \Else 
        \State train $f_\phi$ with standard supervised learning on $\mathcal{D}_l\cup \mathcal{P}$ \Comment{PL training phase}
        \EndIf
        \If{e $\le E_{pt}$ \textbf{and}  e $\% E_{pl} = 0$}
        \State $\mathcal{P} \gets \emptyset$
        \State $\mathcal{P}_{\tau}\gets$ \Call{RebalancedPseudoLabeling}{$\mathcal{D}_u,f_\phi,\tau$}  \Comment{Perform RPL}
        \State $\mathcal{P}_{\gamma}\gets$ \Call{SemanticExplorationClustering}{$\mathcal{D}_u,f_\phi,\gamma$}   \Comment{Perform SEC}
        \State $\mathcal{P} \gets \mathcal{P}_{\tau} \cup \mathcal{P}_{\gamma}$
        \EndIf
        \EndFor
        \State \Return classification model $f_\phi$ 
    \end{algorithmic}
\end{algorithm}

\vspace{-4mm}
\section{Embedding visualization}
\begin{figure*}[h]
	\vspace{-4mm}
	\centering
	\setcounter{subfigure}{0}
	\subfigure[Baseline]{
		\label{fig:subfig:tsne_ce} 
		\includegraphics[width=0.30\linewidth]{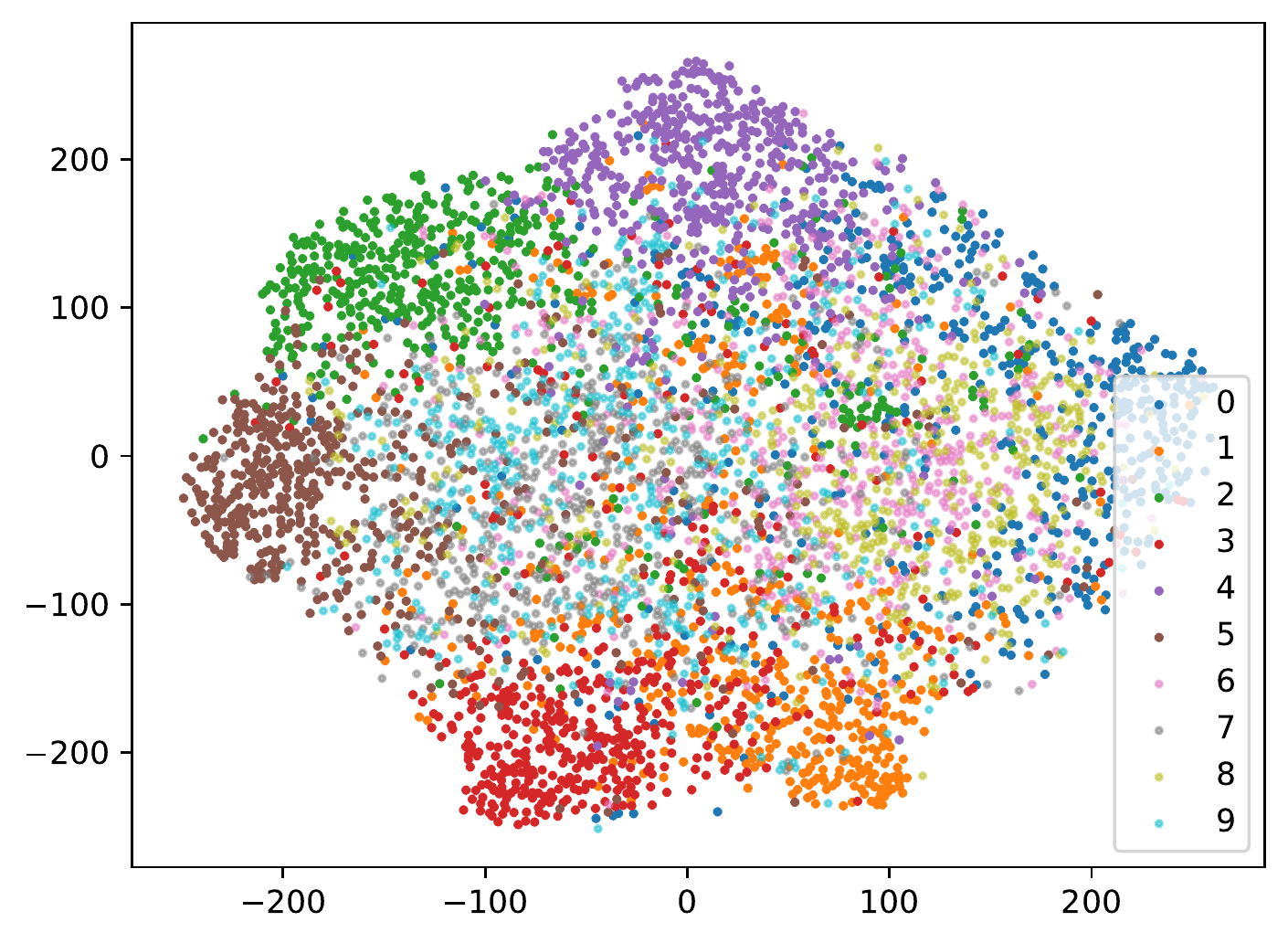}
	}
	\subfigure[PL]{
		\label{fig:subfig:tsne_pl} 
		\includegraphics[width=0.30\linewidth]{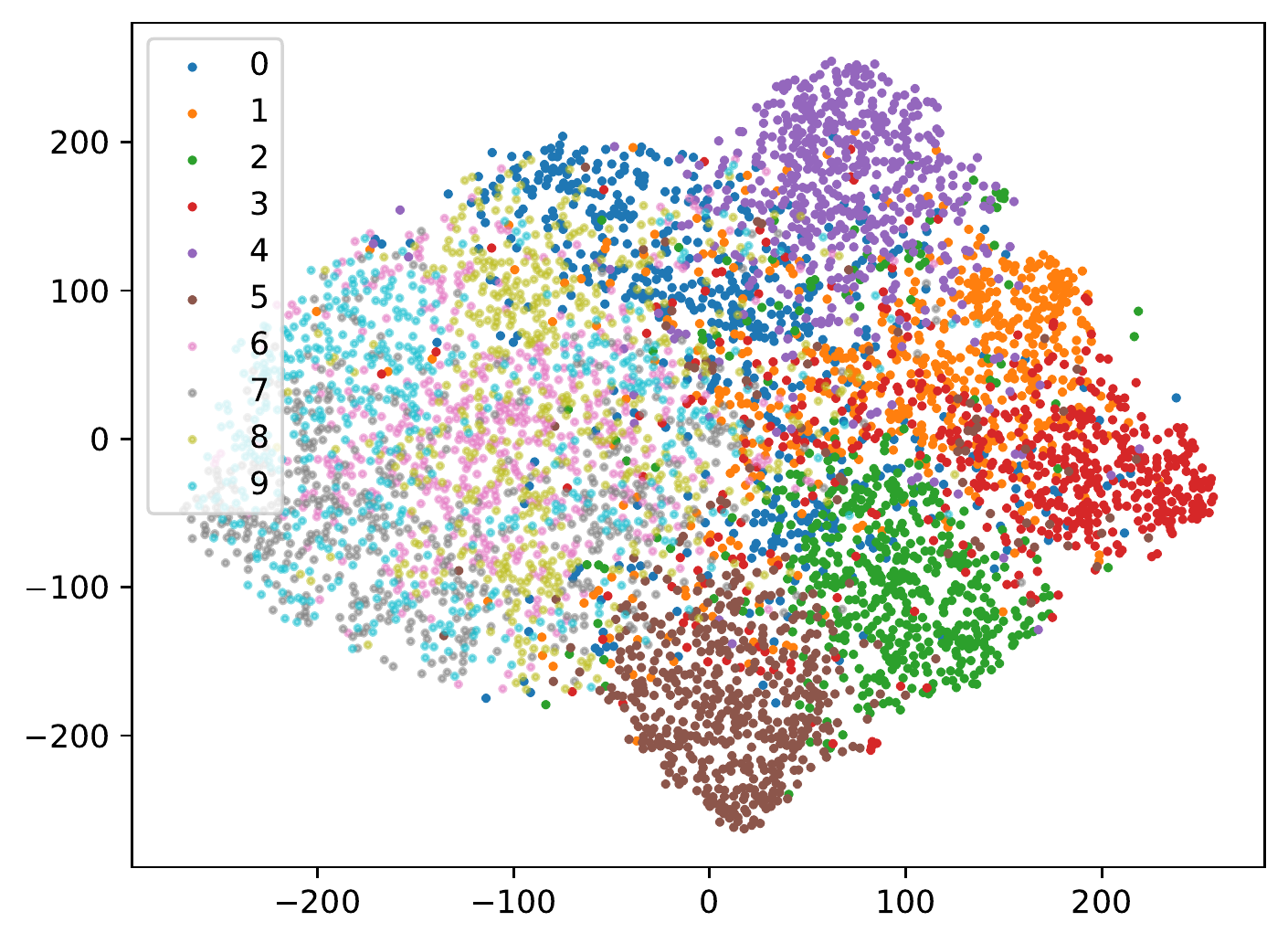}
	}
	\subfigure[$\Upsilon$-Model]{
		\label{fig:subfig:tsne_umodel} 
		\includegraphics[width=0.30\linewidth]{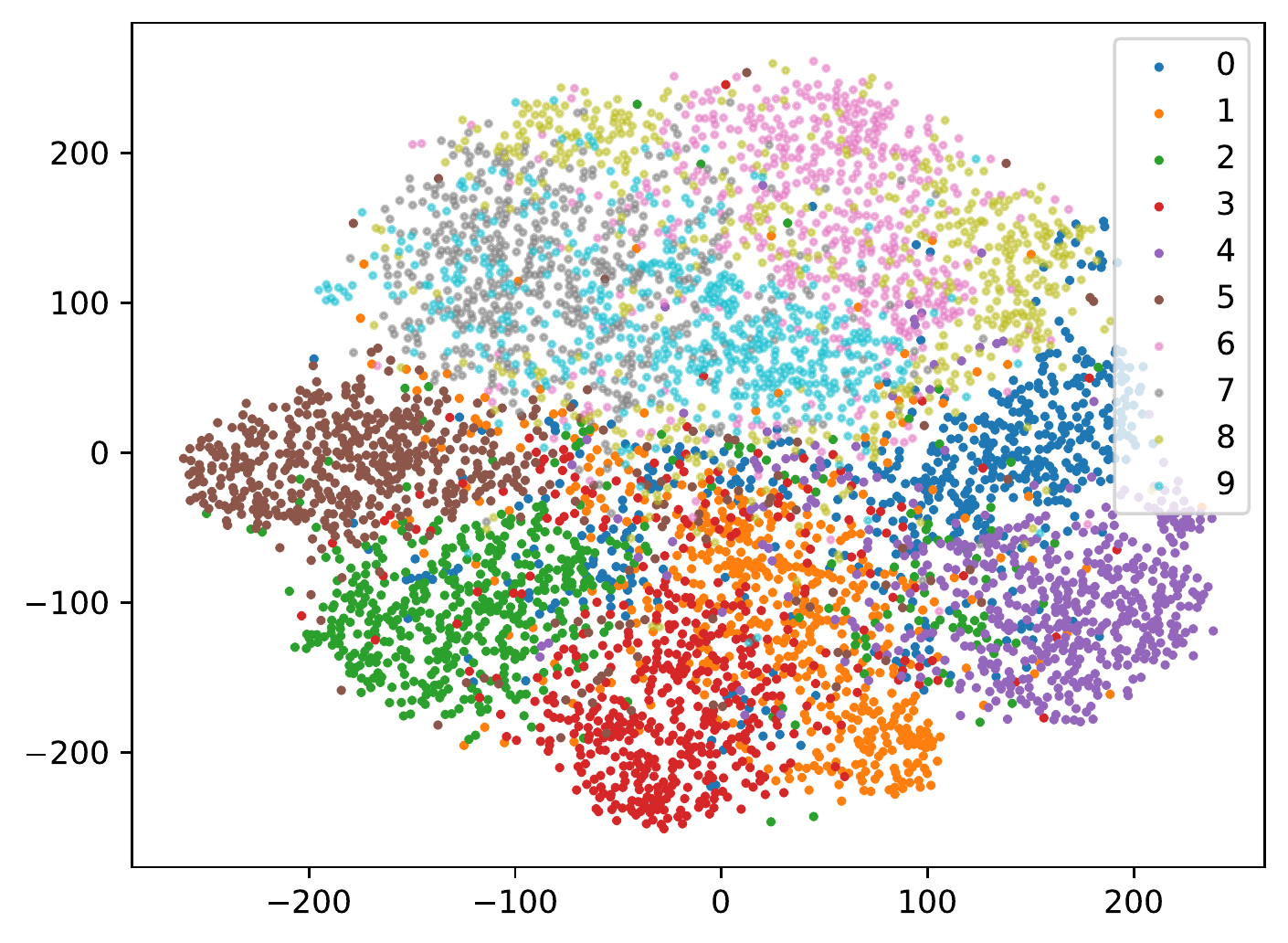}
	}
	\vspace{-4mm}
	\caption{t-SNE visualization of supervised baseline, vanilla PL and $\Upsilon$-Model on CIFAR10's test set. Class 0-5 are ID classes, shown by non-transparent circles. Class 6-9 are OOD classes, represented by semi-transparent circles.}
	\vspace{-4mm}
	\label{fig:tsne} 
\end{figure*}

We visualize the embedding of supervised baseline, vanilla PL and $\upsilon$-Model on CIFAR10's test set with all the classes by t-SNE. 
Figure~\ref{fig:tsne} shows the result. OOD data are mixed with ID data since the supervised baseline does not see unlabeled OOD data. PL mixes OOD data with samples of certain classes (class 0). This is attributed to their pseudo-labels being biased toward this class. Also, we can not clearly distinguish between OOD classes. In contrast, $\Upsilon$-Model can not only make ID data distinguishable but also forms meaningful clusters on OOD data.

\section{Imbalance of Pseudo-Labels}
\label{sec:appendix_imbalance}
\subsection{Metrics}
To demonstrate the imbalance of pseudo-labels on OOD data, we compute two metrics to measure the extent of imbalance.

\begin{itemize}
    \item The KL divergence of pseudo-label distribution $q$ and the uniform distribution $u$: 
    $$kl = KL(q||u) = \sum_{i} q_i \log \frac{q_i}{u_i} $$
    \item The ratio of the `majority class' and `minority class':
    $$r = \frac{\max_i q_i}{\min_i q_i} $$
\end{itemize}

The results on these datasets are displayed in Table \ref{tb:appendix_analysis}.

\begin{table}[htp]
    \caption{Illustration of the extent of imbalance of pseudo-labels on different datasets.}
    \centering
    \begin{tabular}{cccccc}
        \toprule
        \multicolumn{1}{c}{}    & \textbf{C10 (6/4)} & \textbf{C10 (5/5)} & \textbf{SVHN (6/4)} & \textbf{C100 (50/50)} & \textbf{TIN (100/100)} \\ \midrule
        \multicolumn{6}{c}{$kl$}                                                                                            \\\midrule
        \multicolumn{1}{c}{ID} & 0.0131        & 0.0169        & 0.0006     & 0.0771           & 0.0728
                  \\ 
        \multicolumn{1}{c}{OOD} & 0.3636        & 0.0429        & 0.1031     & 0.4177           & 0.4089                 \\ \midrule
        \multicolumn{6}{c}{$r$}                                                                                             \\ \midrule
        \multicolumn{1}{c}{ID}  & 1.5390         & 1.6235        & 1.1047     & 5.1834           & 3.6609              \\ 
        \multicolumn{1}{c}{OOD} & 18.3183       & 2.4878        & 4.4797     & 213.7661         & 187.2124              \\ 
        \bottomrule
    \end{tabular}
    \label{tb:appendix_analysis}
\end{table}

\subsection{Different Selected OOD Classes}
\label{app_subsec:imbalance_ood}
In this section, we show that the imbalance phenomenon is general in \textit{real-world datasets}. We conduct experiments on CIFAR10(6/4), SVHN(6/4), CIFAR100(50/50) and ImageNet100(50/50). The three datasets cover both small-scale and large-scale datasets, cover datasets with both small and large number of classes, and also cover datasets with hierarchical (CIFAR100) and non-hierarchical (SVHN) classes. We also add a dataset with 6 random ID classes from CIFAR10 and 4 random OOD classes from SVHN, denoted as CIFAR10(6)/SVHN(4). This mixed dataset simulates the condition where all the OOD data come from domain largely different from the ID data. For each dataset, we randomly select different set of classes as the ID classes, and the remaining classes as OOD classes except CIFAR10(6)/SVHN(4). A classification model is trained on ID classes and the imbalance ratio on both ID and OOD classes is reported. We repeat for 50 times each dataset. 

\begin{figure}[htp]
    \centering
    \includegraphics[width=0.24\linewidth]{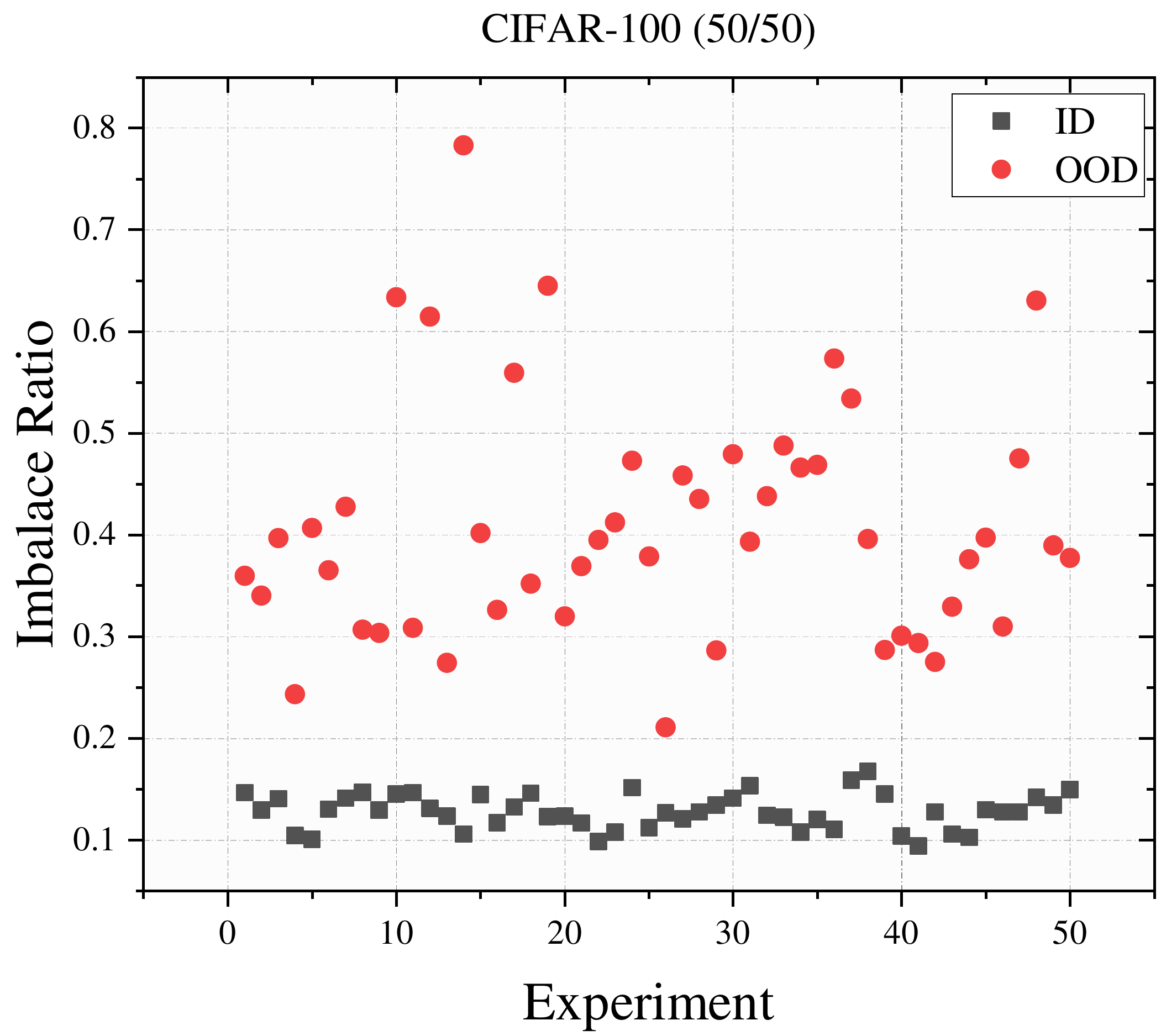}
    \includegraphics[width=0.24\linewidth]{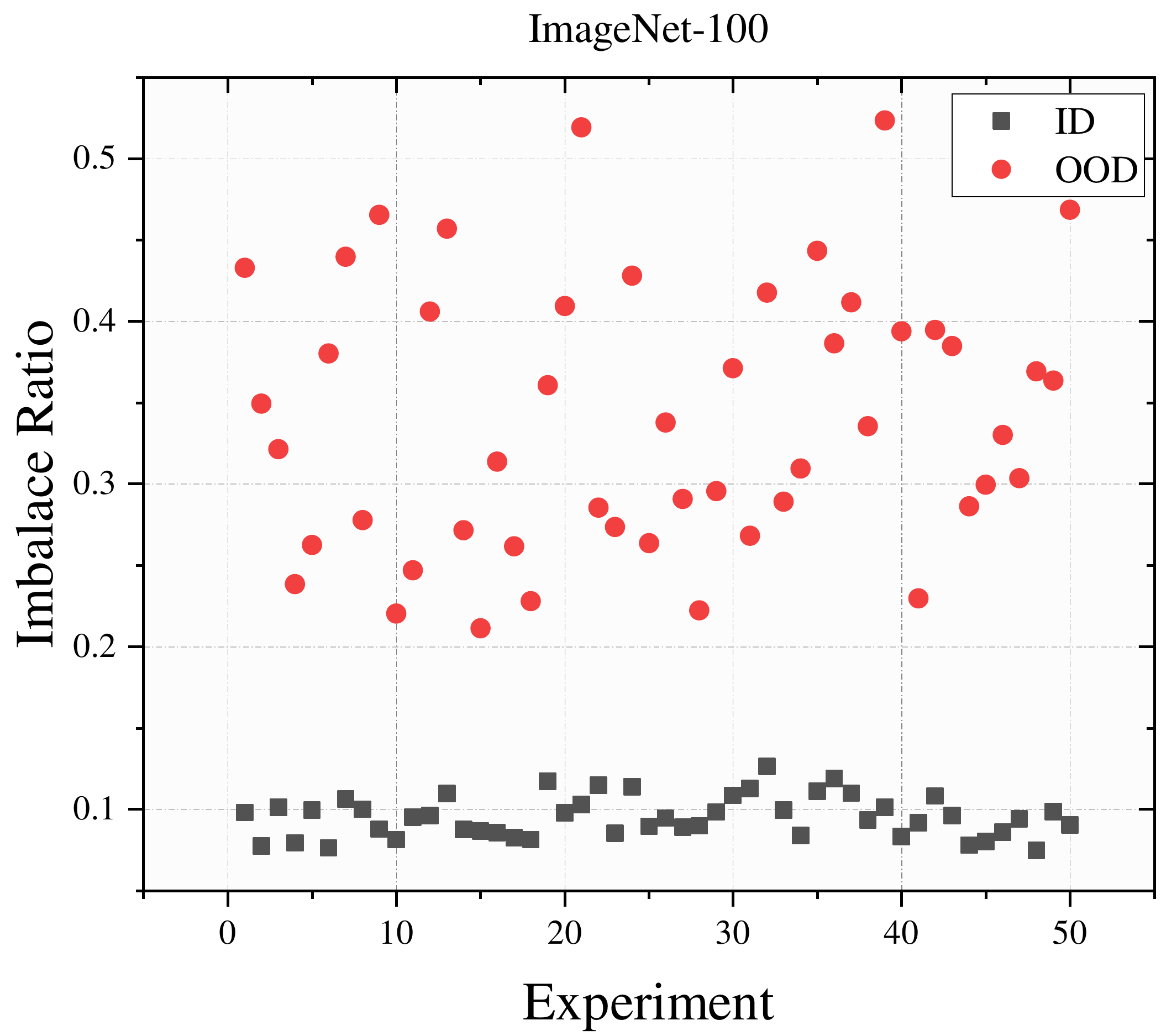}
    \includegraphics[width=0.24\linewidth]{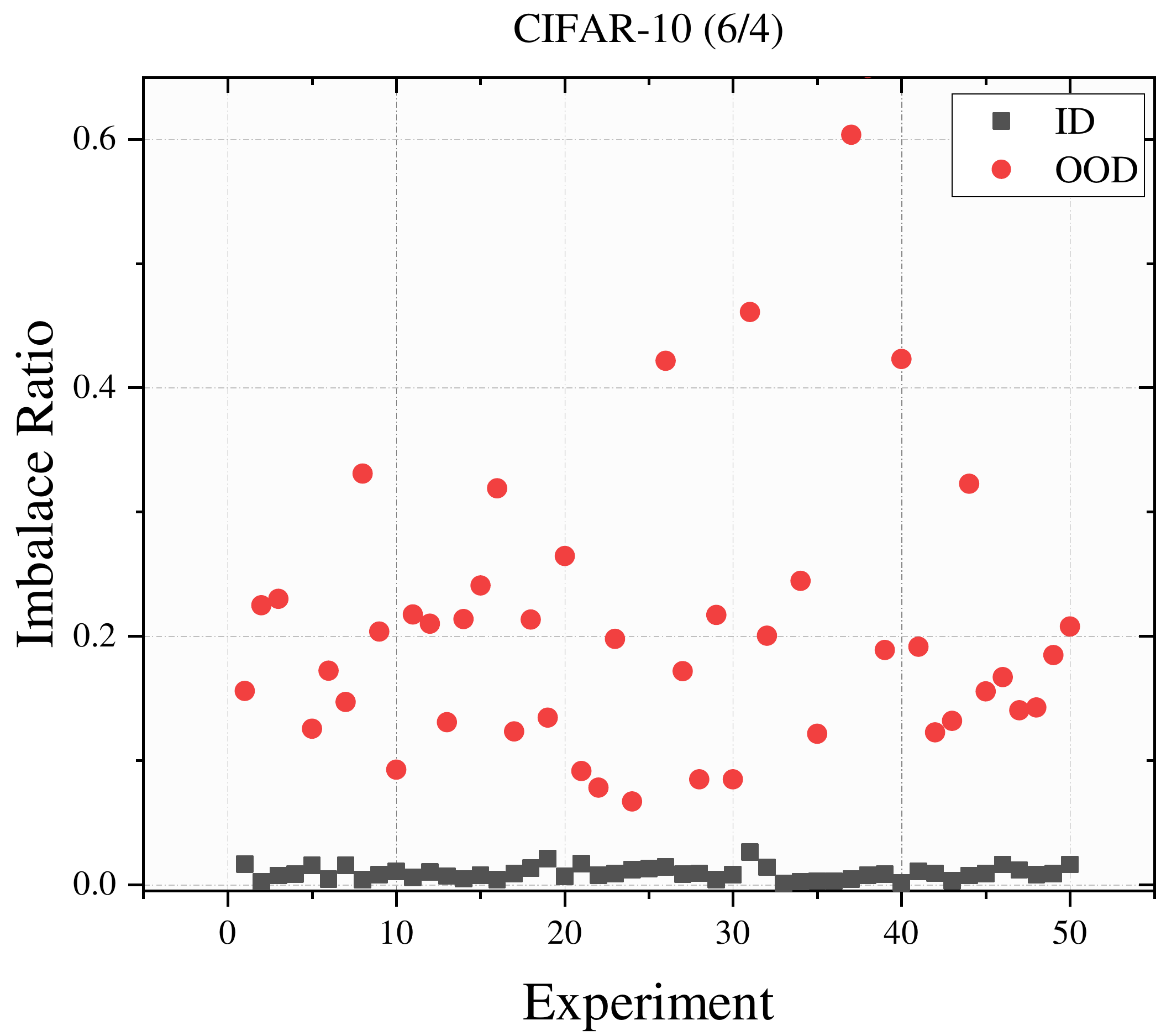}
    \includegraphics[width=0.24\linewidth]{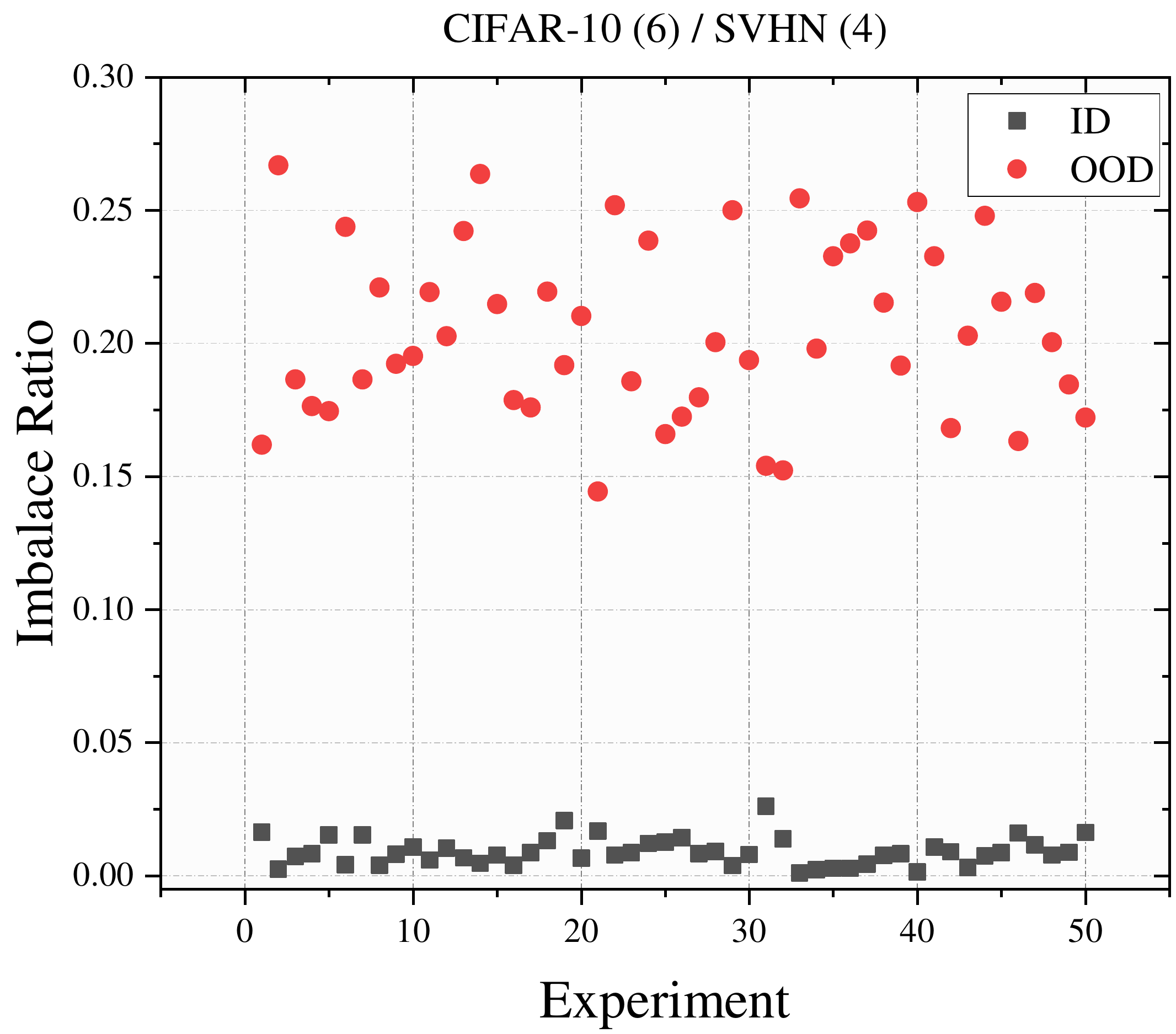}
    \caption{Imbalance ratio on ID and OOD classes with different generations of ID and OOD data. It can be seen that pseudo-labels on ID classes always are always (relatively) balanced. The imbalanced ratios of ID classes concentrate around a certain low value. In contrast, pseudo-labels on OOD data have a much higher imbalanced ratio. Among different OOD settings, the imbalance ratios vary much but all of them lie higher than ratio of ID classes by a large margin. }
    \label{fig:app_imbalance_exp}
\end{figure}

\cref{fig:app_imbalance_exp} display the results on these datasets. We can draw some conclusions from this figure. First, in each trial, the imbalance ratio on ID classes is always lower than OOD by a large margin. It proves that the imbalanced pseudo labels are common on OOD data. Second, imbalance ratios have much lower deviation on ID classes than on OOD classes. It is natural since ID data are selected from the same distribution as the training data, while the OOD data come from a irrelevant domain. Therefore, pseudo-labels on these OOD data appear with high randomization, which means high imbalance ratio and high deviation.

\paragraph{Discussion.} We agree that whether pseudo-labels on OOD data are imbalanced depends on how they are generated. It is always possible to manually pick a set of OOD data that have balanced pseudo-labels. However, we are here to show that for a \textit{non-curated natural dataset}, the imbalance on OOD data is a general phenomenon. With little possibility, the pseudo-labels happen to be balanced on these data. As a result of it, our method is effective in most conditions.



\section{Comparison with FixMatch and MixMatch}
\label{sec:fixmix}
We display the results of FixMatch~\citep{Sohn20Fix} and MixMatch~\citep{Berthelot19Mix}. For a fair comparison, we use the same augmentation for labeled and unlabeled data here. The augmentation 'paper' means what we used in the paper. It is commonly used in class-mismatched settings~\citep{Oliver18Realistic,Guo20Safe}. RandAug~\citep{Cubuk20Rand} is the augmentation used in FixMatch.

\begin{table}[htbp]
    \centering
    \caption{Results of FixMatch, MixMatch, Pseudo-Labeling (PL) and our method. The results are reported on CIFAR-10 (6/4) with different class mismatch ratios.}
    \begin{tabular}{ccccccc}
        \addlinespace
        \toprule
        & Augmentation & 0     & 25\%  & 50\%  & 75\%  & 100\% \\
        \midrule
        MixMatch & paper & 77.88 & 77.15 & 76.83 & 75.9  & 75.85 \\
        PL    & paper & 78.23 & 76.5  & 75.85 & 74.6  & 73.97 \\
        FixMatch & paper & 78.75 & 76.25 & diverge & diverge & diverge \\
        $\Upsilon$-Model (ours) & paper & \textbf{78.98} & \textbf{78.93} &\textbf{77.78} & \textbf{77.9}  & \textbf{78.5} \\
        \midrule
        Baseline & paper & \multicolumn{ 5}{c}{76.21}       \\
        \midrule
        MixMatch & RandAug & 81.98 & 80.08 & 76.75 & 73.45 & 69.33 \\
        PL    & RandAug & 88.30  & 86.80  & 85.61 & 83.72 & 81.62 \\
        FixMatch & RandAug & 87.35 & 84.314 & 82.367 & 78.733 & diverge \\
        $\Upsilon$-Model (ours) & RandAug & \textbf{88.35} & \textbf{86.83} & \textbf{85.91} & \textbf{85.25} & \textbf{84.31} \\
        \midrule
        Baseline & RandAug & \multicolumn{5}{c}{83.75}        \\
        \bottomrule
    \end{tabular}
    \label{tab:fixmix}
\end{table}
Some conclusions can be drawn from~\cref{tab:fixmix}: FixMatch is unstable in the class-mismatched setting while MixMatch is more stable. The former is mainly caused by the imbalance of pseudo-labels. MixMatch's stability may be brought about by its mixup operation. FixMatch and MixMatch all suffer from performance degradation in such a setting. The strong augmentation strategy in FixMatch can bring improvement to our method. It brings improvement to all the methods, but we also emphasize that it also pulls up the baseline.

\section{Hyperparameters on Different Datasets}
\label{sec:appendix_more_datasets}
We compare our method with vanilla PL and the two class-mismatched methods in Section \ref{subsec:exp_cm}. We use the following hyperparameters:
\begin{itemize}
    \item \textbf{CIFAR10 (6/4)}: $\tau=0.95, \gamma = 0.3,E_{pt}=50,E_{pl}=2,K=4$
    \item \textbf{SVHN (6/4)}: $\tau=0.95, \gamma = 0.3,E_{pt}=50,E_{pl}=2,K=4$
    \item \textbf{CIFAR100 (50/50)}:$\tau=0.95, \gamma = 0.18,E_{pt}=50,E_{pl}=2,K=20$
    \item \textbf{Tiny ImageNet (100/100)}: $\tau=0.9, \gamma = 0.15,E_{pt}=50,E_{pl}=2,K=20$
    \item \textbf{ImageNet (50/50)}: $\tau=0.9, \gamma = 0.20,E_{pt}=50,E_{pl}=2,K=20$
\end{itemize}

For \textbf{CIFAR100 (50/50)} , \textbf{Tiny ImageNet (100/100)} and \textbf{ImageNet (50/50)}, we use a weight factor $\lambda$ to trade off the loss on labeled set $\mathcal{D}_l$ and pseudo-labeled set $\mathcal{P}$, which ramps up with function $\lambda=\exp \left(-5 \times\left(1-\min \left(\frac{iter}{40,000}, 1\right)\right)^{2}\right)$, where $iter$ is the number of training steps from $E_{pt}$.

\section{Analysis of Time Complexity}
\label{app:time_complexity}
\begin{figure}[htp]
    \centering
    \includegraphics[width=0.24\linewidth]{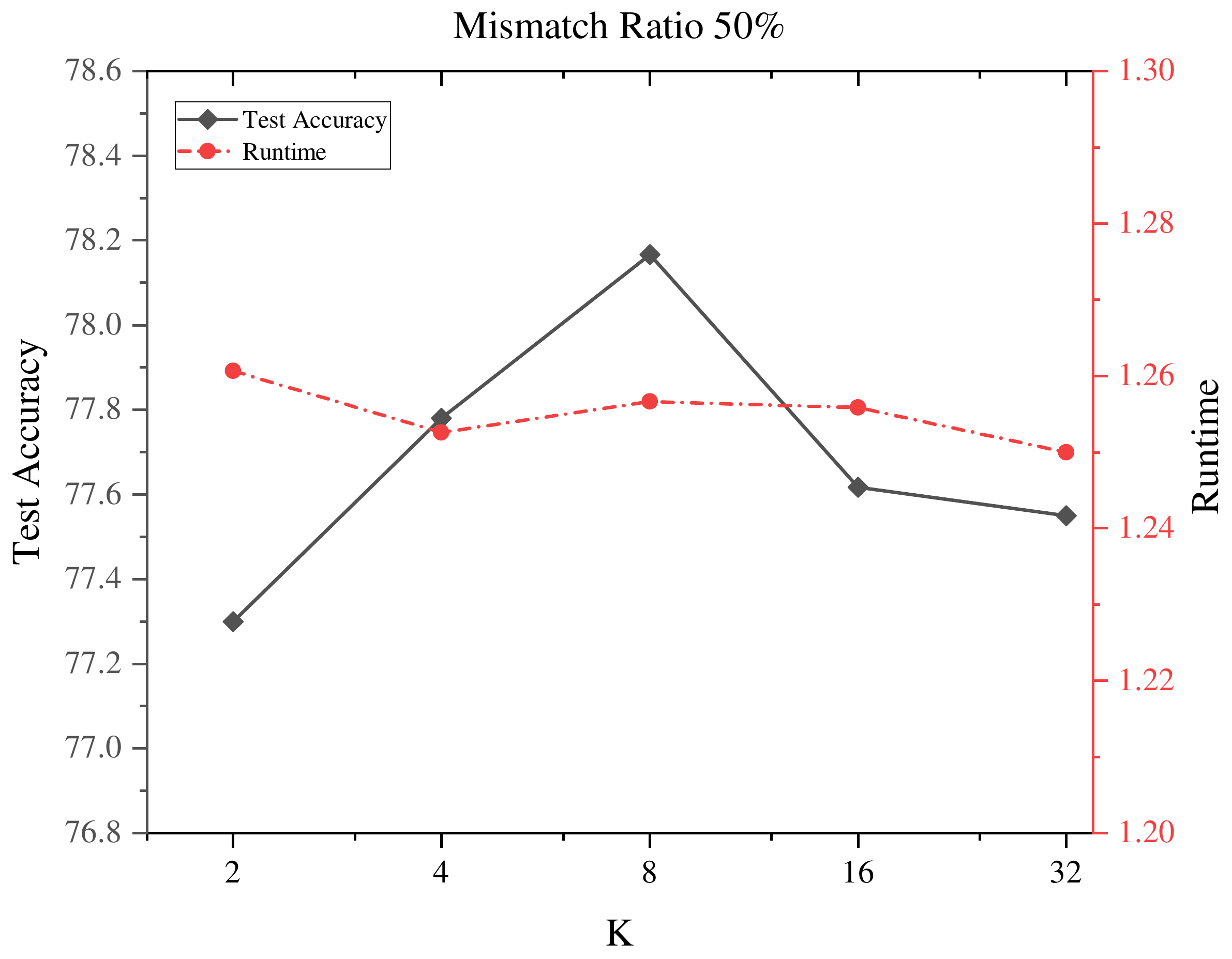}
    \includegraphics[width=0.24\linewidth]{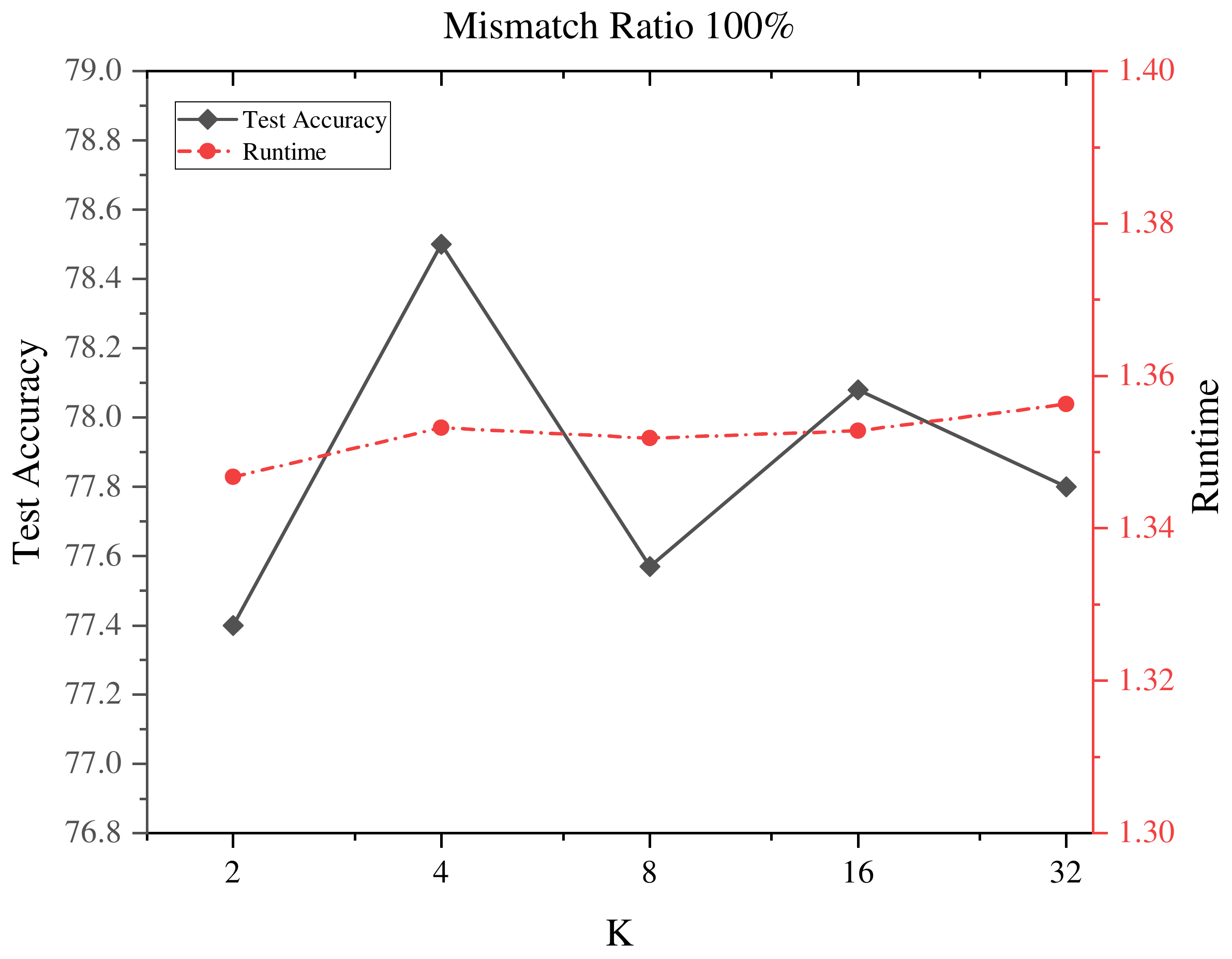}
    \includegraphics[width=0.24\linewidth]{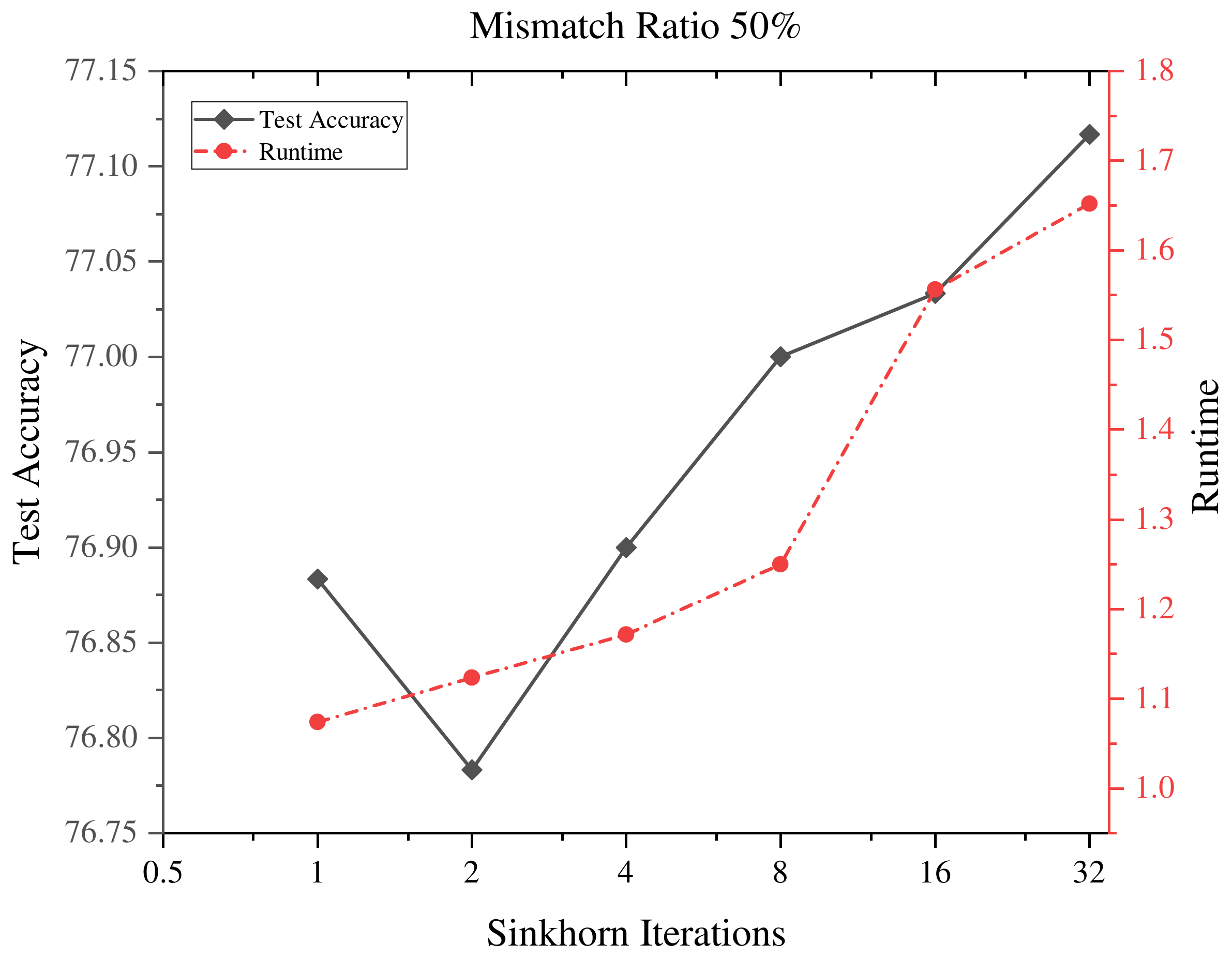}
    \includegraphics[width=0.24\linewidth]{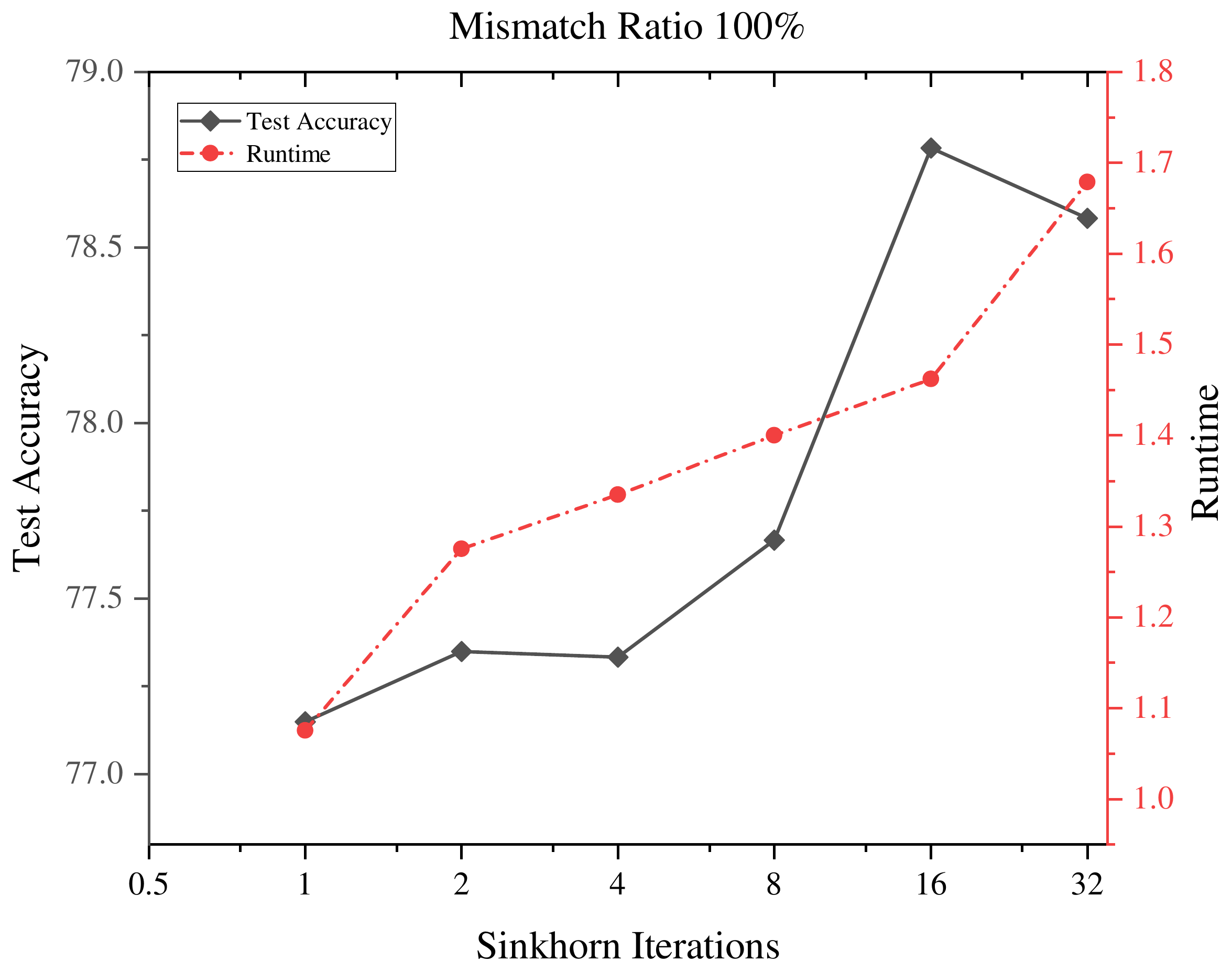}
    \caption{The test accuracy and relative runtime (compare to vanilla Pseudo-Labeling) v.s. the number of clusters $K$ and Sinkhorn iteration. The time complexity is not relevant to $K$, but it affects the test accuracy. Both the accuracy and time complexity increase with the number of iterations. But when the number of iterations increases exponentially, the time increases nearly linearly.}
    \label{fig:time_complex}
\end{figure}

It may be the concern that introducing an extra clustering component causes the $\Upsilon$-Model to be too computationally expensive to be practical. We want to emphasize that it is not the case. We have empirically tested the run time $\Upsilon$-Model compared with vanilla Pseudo-Labeling. We display the comparison results in~\cref{fig:time_complex}. The upper two figures show the test accuracy and relative runtime (compare to vanilla Pseudo-Labeling) with varying numbers of extra classes $K$. It can be shown that the time complexity is almost irrelevant to $K$. The lower two figures show the relationship to the Sinkhorn iterations, which control the quality of clustering~\citep{Cuturi13Sinkhorn,Asano20Self}. It can be seen that both the performance and time complexity increases with the number of iterations. However, we want to note that the X-axis is in log scale. When the number increases exponentially, the time increases nearly linearly. Even though we use 32 iterations, which is 6x the number in~\citet{Asano20Self}, it cost not more than 1.7x time of vanilla Pseudo-Labeling.

Our method is not inefficient as it may seem. First, the OT-based clustering is fast by the Sinkhorn-Knopp algorithm~\citep{Cuturi13Sinkhorn}. Several works using clustering have been proved efficient even on large-scale datasets~\citep{Asano20Self,Caron20Unsupervised}. Our SEC component has similar functionality and setup in practice. Second, we perform clustering periodically. We do clustering every 2 epochs. Compared to the network updating cost, clustering spends an acceptable time.

\section{Discussion about Limitation and Social Impacts}
\label{app:discussion}
 Currently, our model is based on Pseudo-labeling. In future work, we will explore whether other forms of semi-supervised learning methods like the consistency-based method suffer from the same problems. And we will investigate whether the rebalance and semantic explore strategy can also benefit other forms of semi-supervised learning methods. 
 
 This paper explores a way of saving Pseudo-labeling methods from the harm of class-mismatched unlabeled data. The results of this paper can spare the effort of cleaning unlabeled data, which can benefit the development and application of semi-supervised learning. Currently, we do not see a direct negative impact on society.


\end{document}